\documentclass[10pt,twocolumn,letterpaper]{article}

\usepackage[pagenumbers]{cvpr} %
\usepackage{graphicx}
\usepackage{amsmath}
\usepackage{amssymb}
\usepackage{booktabs}
\usepackage{soul}
\usepackage{tabularx}
\usepackage{float}
\usepackage{array}
\usepackage[pagebackref,breaklinks,colorlinks]{hyperref}

\usepackage[capitalize]{cleveref}
\crefname{section}{Sec.}{Secs.}
\Crefname{section}{Section}{Sections}
\Crefname{table}{Table}{Tables}
\crefname{table}{Tab.}{Tabs.}

\def\confName{CVPR}
\def\confYear{2023}

\usepackage{graphics}
\usepackage{overpic}
\usepackage{enumitem} %
\usepackage{overpic} %
\usepackage{color}
\usepackage{paralist}
\usepackage{makecell}
\usepackage{soul} %
\usepackage{pifont}%
\newcommand{\cmark}{\ding{51}}%
\newcommand{\xmark}{\ding{55}}%
\definecolor{turquoise}{cmyk}{0.65,0,0.1,0.3}
\definecolor{purple}{rgb}{0.65,0,0.65}
\definecolor{dark_green}{rgb}{0, 0.5, 0}
\definecolor{orange}{rgb}{0.8, 0.6, 0.2}
\definecolor{red}{rgb}{0.8, 0.2, 0.2}
\definecolor{darkred}{rgb}{0.6, 0.1, 0.05}
\definecolor{blueish}{rgb}{0.0, 0.3, .6}
\definecolor{light_gray}{rgb}{0.7, 0.7, .7}
\definecolor{pink}{rgb}{1, 0, 1}
\definecolor{greyblue}{rgb}{0.25, 0.25, 1}

\newif\ifshowcomments
\showcommentstrue %
\ifshowcomments

\newcommand{\moniker}{PointAvatar}
\newcommand{\x}{\mathbf{\mathbf{x}}}

\newcommand{\xc}{\mathbf{\mathbf{x_{c}}}}

\newcommand{\xd}{\mathbf{\mathbf{x_{d}}}}
\newcommand{\xo}{\mathbf{\mathbf{x_{o}}}}
\newcommand{\MLP}{\textsc{MLP}}
\newcommand{\suppmat}{Supp.~Mat.}

\newcommand{\myparagraph}[1]{\vspace{0.075in}\noindent\textbf{#1}}

\begin{document}
\title{PointAvatar: Deformable Point-based Head Avatars from Videos}
\author{Yufeng Zheng$^{1,2}$
\quad  Wang Yifan$^{3}$
\quad  Gordon Wetzstein$^{3}$
\quad  Michael J. Black$^{2}$
\quad  Otmar Hilliges$^{1}$\\
$^{1}$ETH Zurich, $^{2}$Max Planck Institute for Intelligent Systems, $^{3}$Stanford University
}

\twocolumn[{
	\renewcommand\twocolumn[1][]{#1}%
	\maketitle
 	\begin{center}
    
\includegraphics[trim=0 0 0 0, clip=true, width=\linewidth]{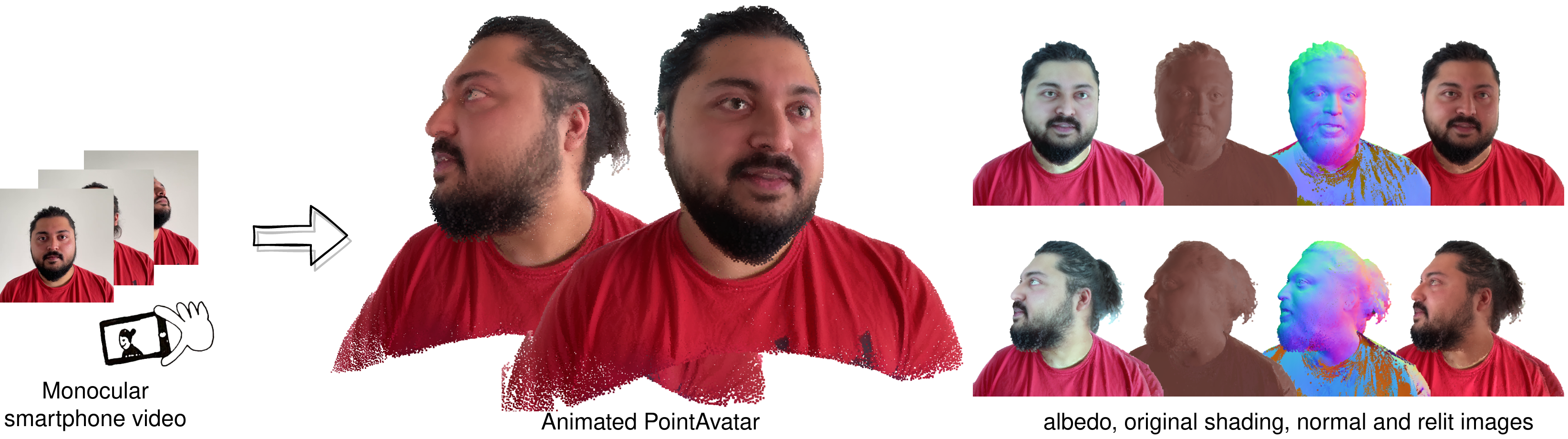}

    \vskip -3mm
    \captionof{figure}{\textbf{PointAvatar} learns lighting-disentangled point-based head avatars from a monocular RGB video captured by a smartphone.}
    \label{fig:teaser}
	\end{center}
}]
\maketitle
\begin{abstract}
The ability to create realistic animatable and relightable head avatars from casual video sequences would open up wide ranging applications in communication and entertainment.
Current methods either build on explicit 3D morphable meshes (3DMM) or exploit neural implicit representations.
The former are limited by fixed topology, while the latter are non-trivial to deform and inefficient to render.
Furthermore, existing approaches entangle lighting in the color estimation, thus they are limited in re-rendering the avatar in new environments.
In contrast, we propose \moniker, a deformable point-based representation that disentangles the source color into intrinsic albedo and normal-dependent shading.
We demonstrate that  \moniker{} bridges the gap between existing mesh- and implicit representations, combining high-quality geometry and appearance with topological flexibility, ease of deformation and rendering efficiency.
We show that our method is able to generate animatable 3D avatars using monocular videos from multiple sources including hand-held smartphones, laptop webcams and internet videos, achieving state-of-the-art quality in challenging cases where previous methods fail, \eg, thin hair strands, while being significantly more efficient in training than competing methods.
\end{abstract}
\let\thefootnote\relax\footnotetext{contact: \href{mailto:yufeng.zheng@inf.ethz.ch}{yufeng.zheng@inf.ethz.ch}}
\let\thefootnote\relax\footnotetext{project page: \href{https://zhengyuf.github.io/PointAvatar/}{https://zhengyuf.github.io/PointAvatar/}}
\section{Introduction}
Personalized 3D avatars are crucial building blocks of the metaverse and beyond. Successful tools for creating avatars should enable easy data capture, efficient computation, and create a photo-realistic, animatable, and relightable 3D representation of the user.
Unfortunately, existing approaches fall short of meeting these requirements.

Recent methods that create 3D avatars from videos either build on 3D morphable models (3DMMs)~\cite{Li_2017_flame, paysan2009bfm} or leverage neural implicit representations~\cite{Mildenhall_2020_nerf, park2019deepsdf, mescheder2019occupancy}.
The former~\cite{Kim_2018_dvp, Grassal_2022_nha, Khakhulin_2022_rome, Buehler_2021_varitex} allows for efficient rasterization and inherently generalizes to unseen deformations,
yet they cannot easily model individuals with eyeglasses or complex hairstyles, as template meshes suffer from a-priori fixed topologies and are limited to surface-like geometries.
Recently, neural implicit representations have also been used to model 3D heads~\cite{Gafni_2021_nerface, Zheng_2022_imavatar, Hong_2022_headnerf, Bergman_2022_gnarf}. While they outperform 3DMM-based methods in capturing hair strands and eyeglasses,
they are significantly less efficient to train and render, since rendering a single pixel requires querying many points along the camera ray.
Moreover, deforming implicit representations in a generalizable manner is non-trivial and existing approaches have to revert to an inefficient root-finding loop, which impacts training and testing time negatively~\cite{Chen_2021_snarf, Zheng_2022_imavatar, wang2022ARAH, li2022tava, jiang2022selfrecon}.

\begin{table}
\resizebox{\linewidth}{!}{%
\begin{tabular}{cccccc}
\toprule
& \makecell{Efficient\\Rendering} & \makecell{Easy\\Animation} & \makecell{Flexible\\Topology} & \makecell{Thin\\Strands} & \makecell{Surface\\Geometry} \\
\midrule
Meshes & \cmark & \cmark & \xmark & \xmark & \cmark \\
Implicit Surfaces & \xmark & \xmark & \cmark & \xmark & \cmark \\
Volumetric NeRF & \xmark & \xmark & \cmark & \cmark & \xmark \\
Points (ours) & \cmark & \cmark & \cmark & \cmark & \cmark \\
\bottomrule
\end{tabular}%
}
\vskip 1mm
\caption{\textbf{PointAvatar} is efficient to render and deform which enables straightforward rendering of full images during training. It can also handles flexible topologies and thin structures and can reconstruct good surface normals in surface-like regions, \eg, skin. }
\label{tab:cmp_table}
\end{table}

To address these issues, we propose \moniker{}, a novel avatar representation that uses point clouds to represent the canonical geometry and learns a continuous deformation field for animation.
Specifically, we optimize an oriented point cloud to represent the geometry of a subject in a canonical space.
For animation, the learned deformation field maps the canonical points to the deformed space with learned blendshapes and skinning weights, given expression and pose parameters of a pretrained 3DMM.
Compared to implicit representations, our point-based representation can be rendered efficiently with a standard differentiable rasterizer.
Moreover, they can be deformed effectively using established %
techniques, \eg, skinning.
Compared to meshes, points are considerably more flexible and versatile. Besides the ability to conform the topology to model accessories such as eyeglasses, they can also represent complex volume-like structures such as fluffy hair.
We summarize the advantanges of our point-based representation in~\cref{tab:cmp_table}.

One strength of our method is the disentanglement of lighting effects. Given a monocular video captured in unconstrained lighting, we disentangle the apparent color into the intrinsic albedo and the normal-dependent shading; see \cref{fig:teaser}.
However, due to the discrete nature of points, accurately computing normals from point clouds is a challenging and costly task~\cite{Huang2009,boulch2012fast,qi2018geonet,lu2020deep}, where the quality can deteriorate rapidly with noise, and insufficient or irregular sampling.
Hence we propose two techniques to
\begin{inparaenum}[(a)]
\item robustly and accurately obtain normals from learned canonical points, and
\item consistently transform the canonical point normals with the non-rigid deformation. %
\end{inparaenum}
For the former, we exploit the low-frequency bias of MLPs~\cite{rahaman2018spectral} and estimate the normals by fitting a smooth signed distance function (SDF) to the points;
for the latter, we leverage the continuity of the deformation mapping and transform the normals analytically using the deformation's Jacobian.
The two techniques lead to high-quality normal estimation, which in turn propagates the rich geometric cues contained in shading to further improve the point geometry. With disentangled albedo and detailed normal directions, \moniker{} can be relit and rendered under novel scenes.

As demonstrated using various videos captured with DSLR, smartphone, laptop cameras, or obtained from the internet, the proposed representation combines the advantages of popular mesh and implicit representations,
and surpasses both in many challenging scenarios.
In summary, our contributions include:
\begin{compactenum}
    \item We propose a novel representation for 3D animatable avatars based on an explicit canonical point cloud and continuous deformation, which shows state-of-the-art photo-realism while being considerably more efficient than existing implicit 3D avatar methods;
    \item We disentangle the RGB color into a pose-agnostic albedo and a pose-dependent shading component;%
    \item We demonstrate the advantage of our methods on a variety of subjects captured through various commodity cameras, showing superior results in challenging cases, \eg, for voluminous curly hair and novel poses with large deformation.
\end{compactenum}
Code is available: \href{https://github.com/zhengyuf/pointavatar}{https://github.com/zhengyuf/pointavatar}

\section{Related Work}
\begin{figure*}[t]
\begin{center}
\includegraphics[trim=0em 0em 0em 0em, clip=true, width=\linewidth]{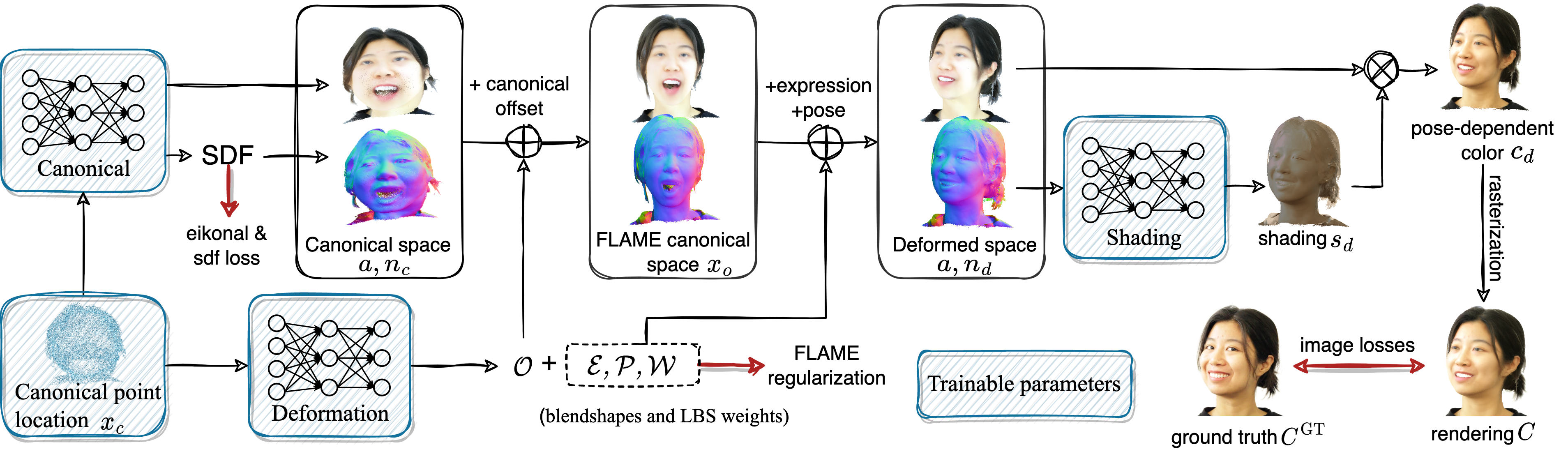}
\end{center}
\caption{
\textbf{Method pipeline.} We model the human head as a learned deformable point cloud consisting of the point locations \(\mathbf{x_c}\), normals \(\mathbf{n_c}\) and albedo \(\mathbf{a}\), which describe the subject's geometry and intrinsic appearance in the canonical space.
To deform points given a target expression and pose in a controllable and interpretable way, we warp \(\x_c\) into the FLAME canonical space via a learned offset \(\mathcal{O}\), and then deform further to the target deformed space by applying the blendshape and skinning using learned personalized blendshape bases (\(\mathcal{E}, \mathcal{P}\)) and skinning weights (\(\mathcal{W}\)).
After obtaining the deformed geometry, a small shading MLP is used to obtain shadings \(\mathbf{s_d}\) from the deformed normals \(\mathbf{n_d}\).
These are multiplied with the albedo colors \(\mathbf{a}\) to produce the shaded colors \(\mathbf{c_{d}}\), which are rendered to images via differentiable rasterization.
\moniker{} leverages a combination of per-pixel and image-based losses to improve photo-realism, while the FLAME regularization term encourages controllable and generalizable animations.}
\label{fig:pipeline}
\end{figure*}
\paragraph{Point clouds for neural rendering.}
Point-based neural rendering is gaining attention thanks to the flexibility and scalability of point representations~\cite{tewari2021advances}.
Aliev \etal~\cite{aliev2020neural} introduce one of the first approaches to point-based neural rendering,
in which each point is augmented with a neural feature.
These features are projected to a set of multi-scale 2D feature maps using a standard point rasterizer, then converted to an RGB image.
Recently, Point-NeRF~\cite{Xu_2022_pointnerf} combines points with volumetric rendering, achieving perceptually high-quality results while being efficient to train. 
Both of the above methods obtain point geometry through off-the-shelf multi-view stereo methods and keep them fixed during optimization, other approaches~\cite{ruckert2022adop,zhang2022differentiable} propose to jointly optimize the point geometry using differentiable point rasterization~\cite{yifan2019differentiable, wiles2020synsin,lassner2021pulsar}.
Our method not only jointly optimizes both the point geometry and colors, but also the deformation of the point cloud through a forward deformation field proposed in~\cite{Zheng_2022_imavatar,Chen_2021_snarf}.

\myparagraph{Head avatars from 2D.}
Learning photo-realistic head avatars from 2D observations is an emerging research topic in the computer vision community. Based on 3DMMs~\cite{paysan2009bfm, Li_2017_flame}, recent works~\cite{Kim_2018_dvp, Grassal_2022_nha} leverage differentiable neural rendering to learn the detailed facial appearance and complete head geometry for a single subject. The idea is extended in~\cite{Buehler_2021_varitex, Khakhulin_2022_rome} to enable generative or one-shot head avatars. 
Another line of work leverages neural implicit representations. NerFace~\cite{Gafni_2021_nerface} extends photo-realistic and flexible neural radiance fields (NeRF)~\cite{Mildenhall_2020_nerf} to model the dynamic head geometry and view-dependent appearance. IMavatar~\cite{Zheng_2022_imavatar} employs neural implicit surfaces~\cite{park2019deepsdf, mescheder2019occupancy, Yariv_2020_multiview, Kellnhofer:2021:nlr} and learns an implicit deformation field for generalizable animation.
Implicit head avatars have been extended to a multi-subject scenario by~\cite{Hong_2022_headnerf, Bergman_2022_gnarf}. To the best of our knowledge, our work is the first to learn deformable point-based head avatars.

\myparagraph{Point-based human avatars.}
PoP~\cite{ma2021pop} learns to model pose-dependent clothing geometry by mapping points from the minimally-clothed SMPL~\cite{matthew2015smpl} surface to the clothing surface and demonstrates impressive geometric quality on various clothing types. %
Point-based clothing representation is extended by \cite{lin2022fite, Ma_3DV_skirt} to mitigate point sparsity issues for long skirts. 
Our work is principally different in two ways: 1) We learn the deformation, geometry, and appearance jointly from scratch, without explicitly relying on a 3DMM template. 2) We learn from monocular videos, whereas previous works~\cite{ma2021pop, lin2022fite, Ma_3DV_skirt} require 3D scans.

\section{Method}
Given a monocular RGB video of a subject performing various expressions and poses, our model jointly learns
\begin{inparaenum}[(1)]
\item a point cloud representing the pose-agnostic geometry and appearance of the subject in a canonical space;
\item a deformation network that transforms the point cloud into new poses using FLAME expression and pose parameters extracted from the RGB frames;
\item a shading network that outputs a per-point shading vector based on the point normals in the deformed space.
\end{inparaenum}
The three components can be jointly optimized by comparing the rendering of the shaded points in the deformed space with the input frames. %
Once trained, we can synthesize new sequences of the same subject under novel poses, expressions and lighting conditions. ~\cref{fig:pipeline} represent an overview of our method.

\subsection{Point-based Canonical Representation}
\begin{figure}[t]
\begin{center}
\includegraphics[trim=0em 0em 0em 0em, clip=true, width=\linewidth]{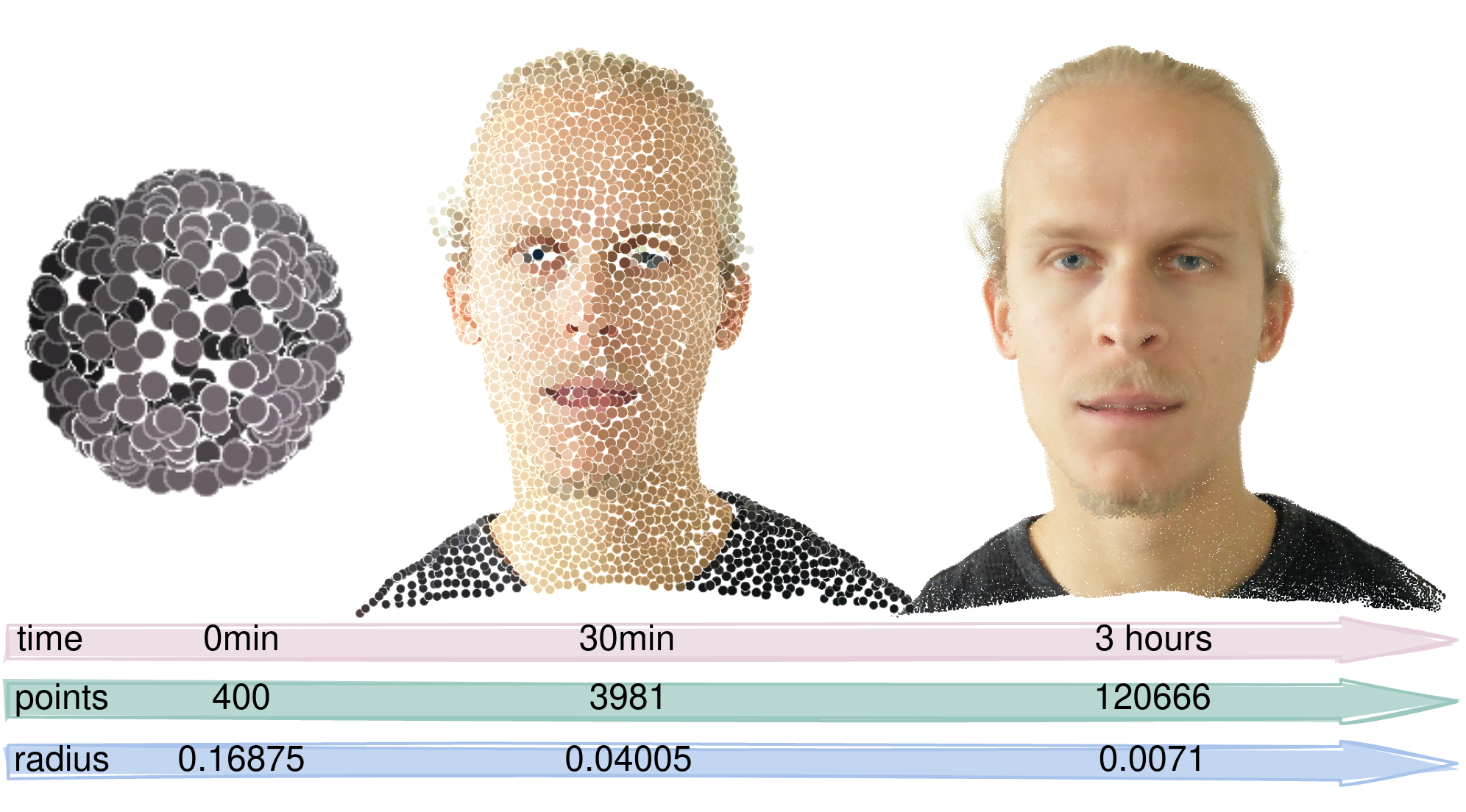}
\end{center}
\caption{
The \textbf{Coarse-to-fine} optimization strategy upsamples points and reduces point radii periodically during training, enableing fast convergence and detailed final reconstruction.}
\label{fig:upsampling}
\end{figure}
Our canonical point representation consists of a set of learnable points $\mathcal{P}_{c}=\{\mathbf{x_c}^i\}$ with $i=\{1, 2, \dots, N\}$,
where \(\xc\in\mathbb{R}^{3}\) denotes the optimizable point locations.
We empirically choose to learn points in an unconstrained canonical space without enforcing them to correspond to a predefined pose, and found that this leads to better geometry quality (implementation explained in \cref{sec:deformation}).

For optimization, we initialize with a sparse point cloud randomly sampled on a sphere and periodically upsample the point cloud while reducing the rendering radii.
This coarse-to-fine optimization scheme enables fast convergence of training because the initial sparse points are efficient to deform and render, and can quickly approximate the coarse shape, whereas denser points in the late training stage lead to good reproduction of details. A schematic of this procedure is provided in \cref{fig:upsampling}.
Furthermore, at the end of every epoch, we prune away invisible points that have not been projected onto any pixels with visibility above a predefined threshold, which further accelerates training.

Our method disentangles the point colors into a pose-agnostic albedo component and a pose-dependent shading component (explained later in \cref{sec:shading}), where the shading component is inferred from the point normals in the deformed space.
We first discuss how to compute the normals and the albedo in canonical space, followed by the learned transformation of point locations and normals into the deformed space (see \cref{sec:deformation}).

\paragraph{Canonical normals.}
Point normals are local geometric descriptors that can be estimated from neighboring point locations.
However, estimating normals in this way yields poor results, especially when the points are sparse, noisy and irregularly sampled, as is the case during early training.
To alleviate this issue, we propose to estimate the point normals from an SDF defined by the canonical points. Normals are then defined as the spatial gradient of the SDF:
\begin{equation}
    \mathbf{n_c} = \nabla_{\mathbf{x_c}}\textsc{SDF}\left(\mathbf{x_c}\right).
\end{equation}

Specifically, we represent the SDF with an MLP and fit the zero-level set to the canonical point locations using a data term and an Eikonal regularizer, defined as:
\begin{equation}
\mathcal{L}_{\textrm{sdf}} = \left\Vert \textsc{SDF}\left(\mathbf{x_c}\right)\right\Vert^2 \,\textrm{and}\,
\mathcal{L}_{\textrm{eik}} = \left(\left\Vert \nabla_{\mathbf{x_e}}\textsc{SDF}\left(\mathbf{x_e}\right) \right\Vert - 1\right)^2,
\end{equation}
where the Eikonal samples $\mathbf{x_e}$ consist of the original and perturbed point locations~\cite{Gropp_2020_igr}. We update the SDF using current point locations at each training step. %

\paragraph{Canonical albedo.}
\label{sec:method_canonical}
We use an MLP to map the point locations \(\xc\) to the albedo colors \(\mathbf{a} \in \mathbb{R}^3\), similar to \cite{caselles2022sira}.
Compared with directly modeling the albedo of each point as an individual per-point feature,
the inductive bias of MLPs automatically enforces a local smoothness prior on albedo colors~\cite{tappen2002recovering,grosse2009ground,barron2011high,barron2012shape}.
For efficiency, we use a shared MLP to compute the canonical normals and albedo in practice, \ie,
\begin{equation}
\left[\textsc{SDF}\left( \xc \right); \mathbf{a}\right]= \MLP_{\textrm{c}}\left( \xc \right).\label{eq:canonical_sdf}
\end{equation}

\subsection{Point Deformation}\label{sec:deformation}
In order to achieve controllable animations, \moniker{} is deformed using the same expression and pose parameters as FLAME~\cite{Li_2017_flame}, a parametric head model learned from 4D scans.
Our deformation takes a two-step approach shown in \cref{fig:pipeline}.
In the first stage, we warp the canonical points \(\mathbf{x_{c}}\) to its location in an intermediate pose \(\mathbf{x_o}\), which corresponds to a predefined mouth-opened canonical pose of the FLAME template.
In the second stage, we use the target FLAME expression and pose parameters to transform \(\mathbf{x_o}\) to the deformed space based on learned blendshapes and skinning weights. %
Our experiments (see \cref{sec:ablation_canonical}) empirically show that the proposed two-staged deformation helps to avoid bad local minima during optimization and yields more accurate geometries.

To improve expression fidelity and to account for accessories such as eyeglasses, we learn personalized deformation blendshapes and skinning weights, similar to IMavatar~\cite{Zheng_2022_imavatar}. %
Specifically, a coordinate-based MLP is used to map each canonical point \(\xc\) to
\begin{inparaenum}[(1)]
\item an offset \(\mathcal{O} \in \mathbb{R}^{3}\), which translates \(\xc\) to its corresponding location in the FLAME canonical space \(\xo\);
\item \(n_{e}\) expression blendshapes and \(n_{p}\) pose blendshapes, denoted as \(\mathcal{E}\in\mathbb{R}^{n_{e}\times 3}\) and \(\mathcal{P}\in\mathbb{R}^{n_{p}\times 9\times 3}\);
\item LBS weights \(\mathcal{W}\in\mathbb{R}^{n_{j}}\) associated to the \(n_{j}\) bones.
\end{inparaenum}
The point location in deformed space \(x_d\) is then computed as:
\begin{gather}
\xo = \xc + \mathcal{O}\label{eq:xo}\\
\xd = \textsc{LBS}\left( \xo + \textsc{B}_{P}\left(\mathbf{\theta}; \mathcal{P} \right) + \textsc{B}_{E}\left( \mathbf{\psi}; \mathcal{E} \right), \textsc{J}\left(\mathbf{\psi} \right), \mathbf{\theta}, \mathcal{W} \right),\label{eq:x_d}
\end{gather}
where \textsc{LBS} and \textsc{J} define the standard skinning function and the joint regressor defined in FLAME, and \(\textsc{B}_{P}\) and \(\textsc{B}_{E}\) denote the linear combination of blendshapes, which outputs the additive pose and expression offsets from the animation coefficients $\theta$ and $\psi$ and the blendshape bases $\mathcal{P}$ and $\mathcal{E}$.
Despite the similarity of this formulation to IMavatar~\cite{Zheng_2022_imavatar}, our forward deformation mapping only needs to be applied once in order to map canonical point locations \(x_c\) to the deformed space, and therefore does not need the computationally costly correspondence search of IMavatar~\cite{Zheng_2022_imavatar}.

\paragraph{Normal deformation.}\label{sec:normal_deformation}
The deformation mapping defined in \cref{eq:x_d} is differentiable \wrt the input point locations.
This allows us to transform canonical normals analytically with the inverse
of the deformation Jacobian

\begin{equation}
\mathbf{n_d} = l\mathbf{n_c}\left(\frac{\mathrm{d} \mathbf{x_d}}{\mathrm{d} \mathbf{x_c}}\right)^{-1},
\label{eq:n_d}
\end{equation}
where \(l\) is a normalizing scalar to ensure the normals are of unit length.
The formulation can be obtained via Taylor's theorem. Please check \suppmat{} for the detailed proof.
\subsection{Point Color}
\label{sec:shading}
The color of each point in the deformed space \(\mathbf{c_{d}}\) is obtained by multiplying the intrinsic albedo \(\mathbf{a} \in \mathbb{R}^3\) with the pose-dependent lighting effects which we refer to as shading \(\mathbf{s_{d}} \in \mathbb{R}^3\): 
\begin{align}
\mathbf{c_{d}} &= \mathbf{a} \circ \mathbf{s_{d}} ,\label{eq:shading}
\end{align}
where \(\circ\) denotes the Hadamard product.
As we assume our input video is captured under a fixed lighting condition and camera position, we model the shading as a function of the deformed point normals \(\mathbf{n_{d}}\).
In practice, we approximate their relation using a shallow MLP:
\begin{equation}
    \mathbf{s_{d}} = \textsc{MLP}_{s}\left( \mathbf{n_{d}} \right)\label{eq:shading_mlp}
\end{equation}
By conditioning the albedo only on the canonical locations (see Sec.~\ref{sec:method_canonical}) and the shading only on the normal directions, our formulation achieves \emph{unsupervised} albedo disentanglement. 
Despite that our lighting model is simple and minimally constrained, our method can perform rudimentary relighting by changing the shading component (see \cref{sec:relighting}).

\subsection{Differentiable Point Rendering}\label{sec:DPR}
One advantage of our point-based representation is that it can be rendered efficiently using rasterization.
We adopt PyTorch3D's~\cite{Ravi_2020_pytorch3d} differentiable point renderer.
It splats each point as a 2D circle with a uniform radius, arranges them in z-buffers, and finally composites the splats using alpha compositing.
The alpha values are calculated as $\alpha = 1 - d^2/r^2$, where $d$ is the distance from the point center to the pixel center and $r$ represents the splatting radius.
Then, the transmittance values are calculated as $T_i = \prod_{k=1}^{i-1} (1 - \alpha_k)$, where \(i\) denotes the index of the sorted points in the z-buffer.
The point color values are integrated to produces the final pixel color: \begin{equation}
    c_{\textrm{pix}} = \sum_i \alpha_i T_i c_{\textrm{d}, i}.\label{eq:pixel_color}
\end{equation}

\subsection{Training Objectives}
The efficiency of point-based rendering allows us to render the whole image at each training step. This makes it possible to apply perceptual losses on the whole image, whereas implicit-based avatar methods are often limited to per-pixel objectives.
Specifically, we adopt VGG feature loss~\cite{Johnson2016Perceptual}, defined as
\begin{equation}
 \mathcal{L}_{\textrm{vgg}}\left( \mathbf{C} \right)=\left\Vert \textsc{F}_{\text{vgg}}(\mathbf{C})-\textsc{F}_{\text{vgg}}(\mathbf{C}^\textrm{GT})\right\Vert,\label{eq:vgg_loss}
\end{equation}
where \(\mathbf{C}\) and \(\mathbf{C}^\textrm{GT}\) denote the predicted and ground truth image, and $\textsc{F}_{\textrm{vgg}}(\cdot)$ calculates the features from the first four layers of a pre-trained VGG~\cite{Simonyan_2014_vgg} network.

In addition to \(\mathcal{L}_{\textrm{vgg}}\), we follow prior work and adopt the losses of IMavatar~\cite{Zheng_2022_imavatar}:\\
\resizebox{.9\linewidth}{!}{
  \begin{minipage}{\linewidth}
\begin{gather*}
\mathcal{L}_{\textrm{RGB}} = \left\Vert \mathbf{C} - \mathbf{C}^\textrm{GT} \right\Vert, \\ %
\mathcal{L}_{\textrm{flame}} = \frac{1}{N}\sum_{i=1}^{N}(\lambda_e\|\mathcal{E}_{i} - \widehat{\mathcal{E}}_{i}\|_2 +\lambda_p \|\mathcal{P}_{i}-\widehat{\mathcal{P}}_{i}\|_2 + \lambda_w\|\mathcal{W}_{i}-\widehat{\mathcal{W}}_{i}\|_2), \\
\mathcal{L}_{\textrm{mask}} = \left\Vert \mathbf{M} - \mathbf{M}^\textrm{GT} \right\Vert.\\ %
\end{gather*}
\end{minipage}
}
Here, \(\mathbf{M}\) and \(\mathbf{M}^{\textrm{GT}}\) denote the predicted and ground truth head mask,  \(\widehat{\mathcal{E}}, \widehat{\mathcal{P}}\) and \(\widehat{\mathcal{W}}\) are pseudo ground truth defined by the nearest FLAME vertex.
The mask prediction at each pixel is obtained by \(m_\textrm{pix}=\sum_i \alpha_i T_i\), and the ground truth mask obtained using an off-the-shelf foreground estimator~\cite{MODNet},

Our total loss is
\begin{equation}
\mathcal{L} = \lambda_{\textrm{rgb}} \mathcal{L}_{\textrm{RGB}} + \lambda_{\textrm{mask}}\mathcal{L}_{\textrm{mask}} + \lambda_{\textrm{flame}}\mathcal{L}_{\textrm{flame}} + \lambda_{\textrm{vgg}}\mathcal{L}_{\textrm{vgg}}.\label{eq:total_loss}
\end{equation}
The loss weights can be found in \suppmat{} together with other implementation details.

\section{Experiments}
\paragraph{Datasets.}
We compare our approach with state-of-the-art (SOTA) methods on 2 subjects from IMavatar~\cite{Zheng_2022_imavatar} and 2 subjects from NerFace~\cite{Gafni_2021_nerface}. Additionally, we evaluate different methods on 1 subject collected from the internet, 4 subjects captured with hand-held smartphones, and 1 subject from a laptop webcam. These settings pose new challenges to avatar methods due to limited head pose variation, automatic exposure adjustment or low image resolution. In Supp.~Mat., we evaluate reconstructed point geometry on a synthetic dataset rendered from the MakeHuman project~\cite{Briceno_2019_makehuman}. For all subjects, we use the same face-tracking results for all comparing methods. 

\myparagraph{Baselines.}
We compare our method with three SOTA baselines, including 
\begin{inparaenum}[(1)]
\item NerFace~\cite{Gafni_2021_nerface}, which leverages dynamic neural radiance fields~\cite{Mildenhall_2020_nerf}, 
\item neural head avatar (NHA)~\cite{Grassal_2022_nha} based on 3D morphable mesh models~\cite{Li_2017_flame} 
\item and IMavatar~\cite{Zheng_2022_imavatar}, which builds on neural implicit surfaces~\cite{Yariv_2020_multiview} and learnable blendshapes.
\end{inparaenum} 
Together with our method, which represents the head geometry with deformable point clouds, our experiments reveal the strength and weaknesses of each geometry representation in the scenario of head avatar reconstruction. Through our experiments, we demonstrate the efficiency, flexibility and photo-realism achieved by point-based avatars.

\subsection{Comparison with SOTA Methods}

\begin{table}
\resizebox{\linewidth}{!}{
    \begin{tabular}{lcccc}
    \toprule
     & L1 $\downarrow$ & LPIPS $\downarrow$& SSIM $\uparrow$& PSNR $\uparrow$\\
    \midrule
    IMavatar~\cite{Zheng_2022_imavatar} & 0.033; 0.050 & 0.178; 0.261 & 0.874; 0.770 & 22.4; 18.7\\
    NerFace~\cite{Gafni_2021_nerface} & 0.030; 0.045 & 0.126; 0.187 & 0.877; 0.782 & 22.7; 19.6\\
    Ours & \textbf{0.021}; \textbf{0.036}&\textbf{0.094}; \textbf{0.145} &\textbf{0.899}; \textbf{0.802} &\textbf{26.6}; \textbf{22.3}\\
    \midrule
    NHA~\cite{Grassal_2022_nha} & 0.022;0.029 & 0.086; 0.123 & 0.890; 0.837	& 25.7;  21.6\\
    Ours (no cloth)& \textbf{0.017};\textbf{ 0.021} & \textbf{0.077}; \textbf{0.100} &\textbf{0.912}; \textbf{0.863} & \textbf{28.6}; \textbf{25.8}\\
    \bottomrule
    \end{tabular}
}
\caption{\textbf{Quantitative comparison.} The first and second number in each cell represent scores for lab-capture sequences (IMavatar and NerFace datasets) and casual videos (smartphone, webcam and internet videos), respectively.
\moniker{} outperforms implicit-based IMavatar and NerFace by a large margin in perceptual quality reflected by LPIPS~\cite{zhang18lpips}. Compared to 3DMM-based NHA, our method not only generates complete avatars with shoulders and clothing, but also performs better for the head region. }

\label{tab:sota_real}
\end{table}
\begin{figure*}
\begin{center}
\setlength\tabcolsep{1pt}
\setlength\extrarowheight{-10pt}
\newcommand{\myheight}{4cm}
\begin{tabularx}{\textwidth}{cccccc}
\includegraphics[height=\myheight]{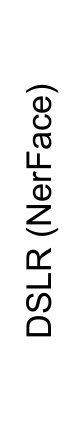}
&\includegraphics[trim=0em 0em 0em 1em, clip=true, height=\myheight]{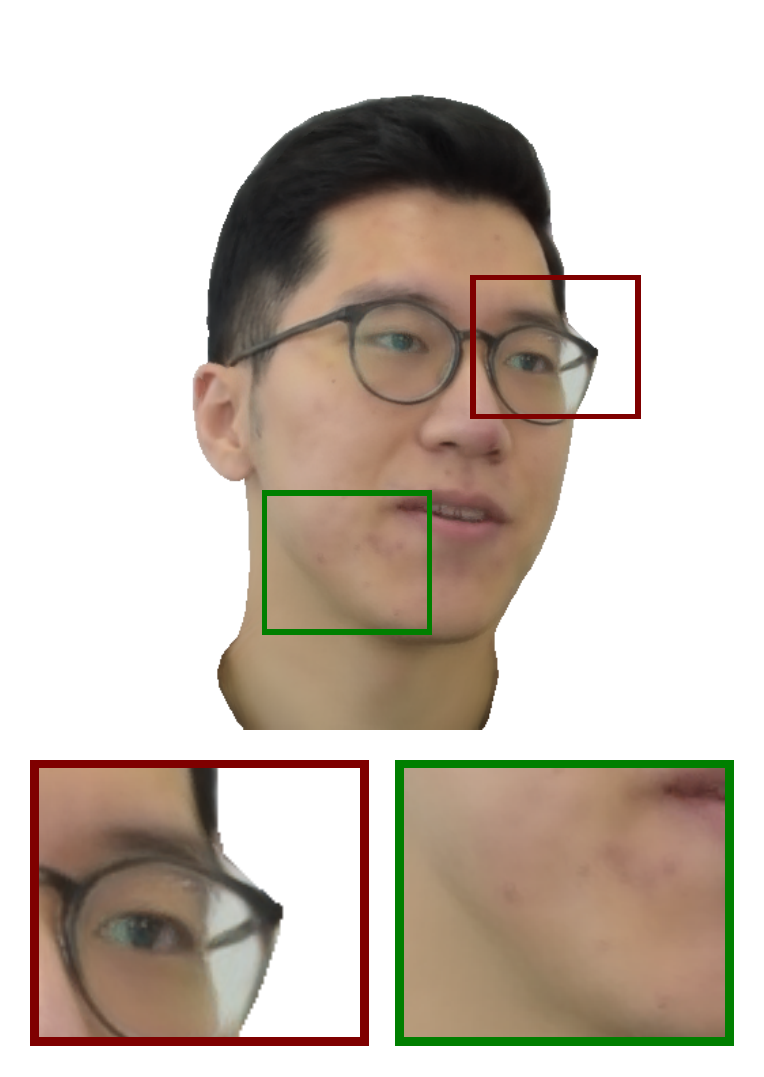}
&\includegraphics[trim=0em 0em 0em 1em, clip=true, height=\myheight]{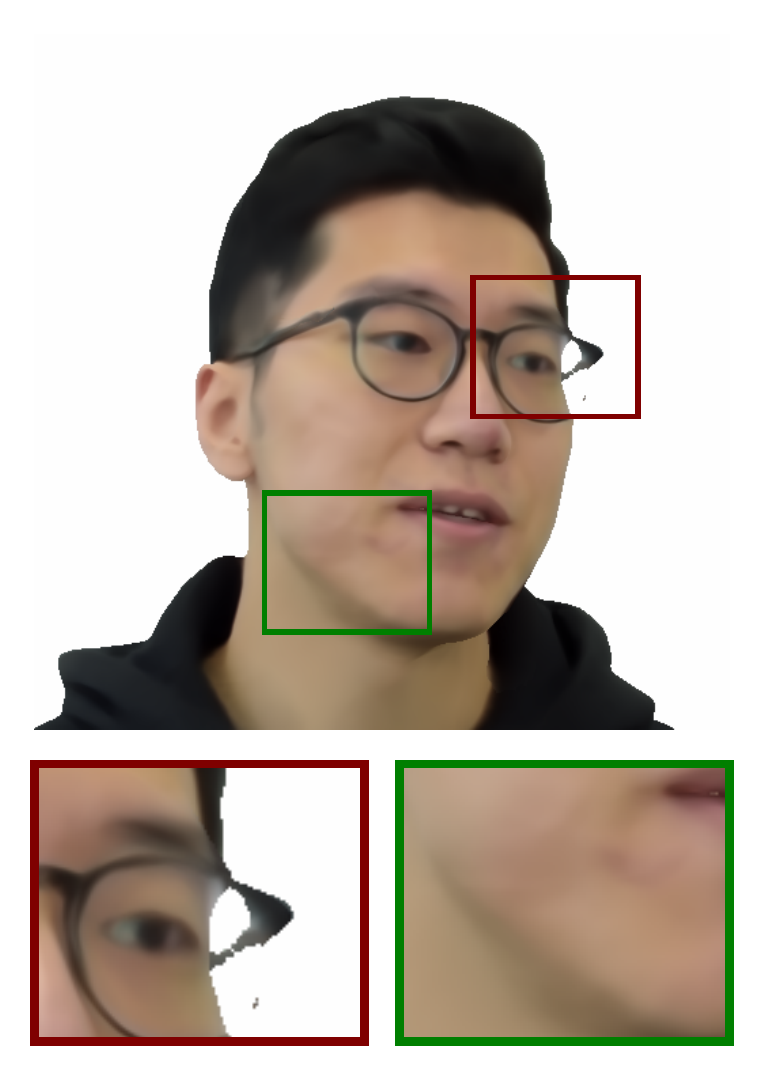}
&\includegraphics[trim=0em 0em 0em 1em, clip=true, height=\myheight]{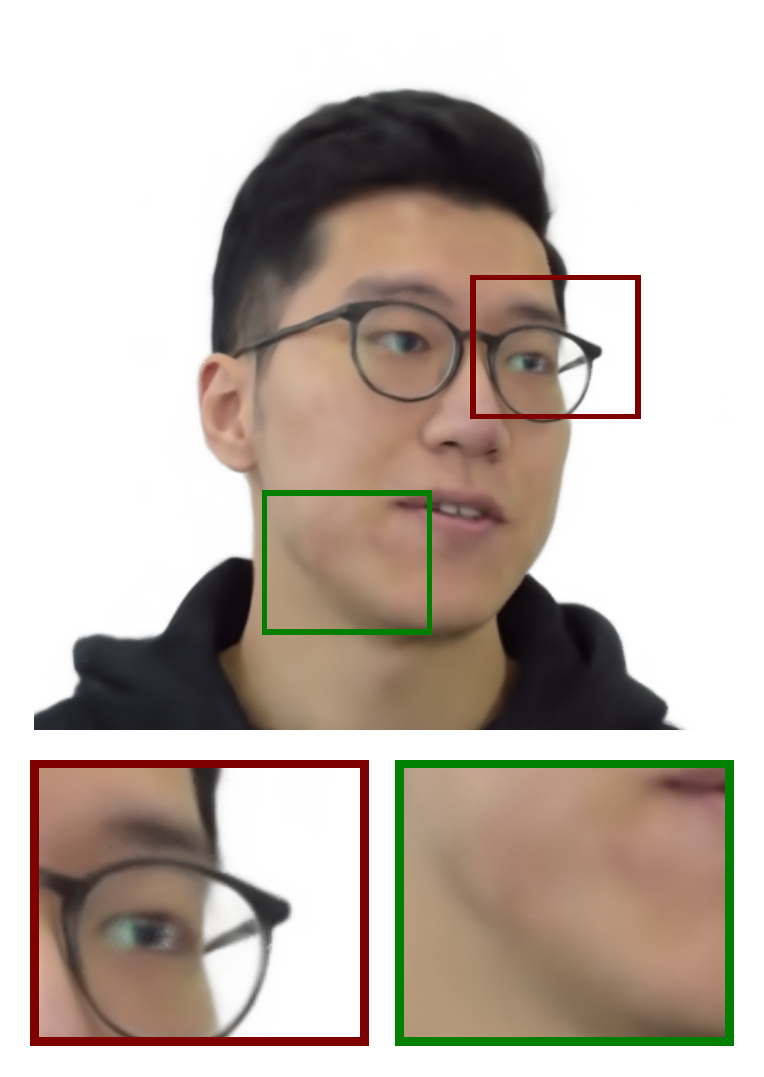}
&\includegraphics[trim=0em 0em 0em 1em, clip=true, height=\myheight]{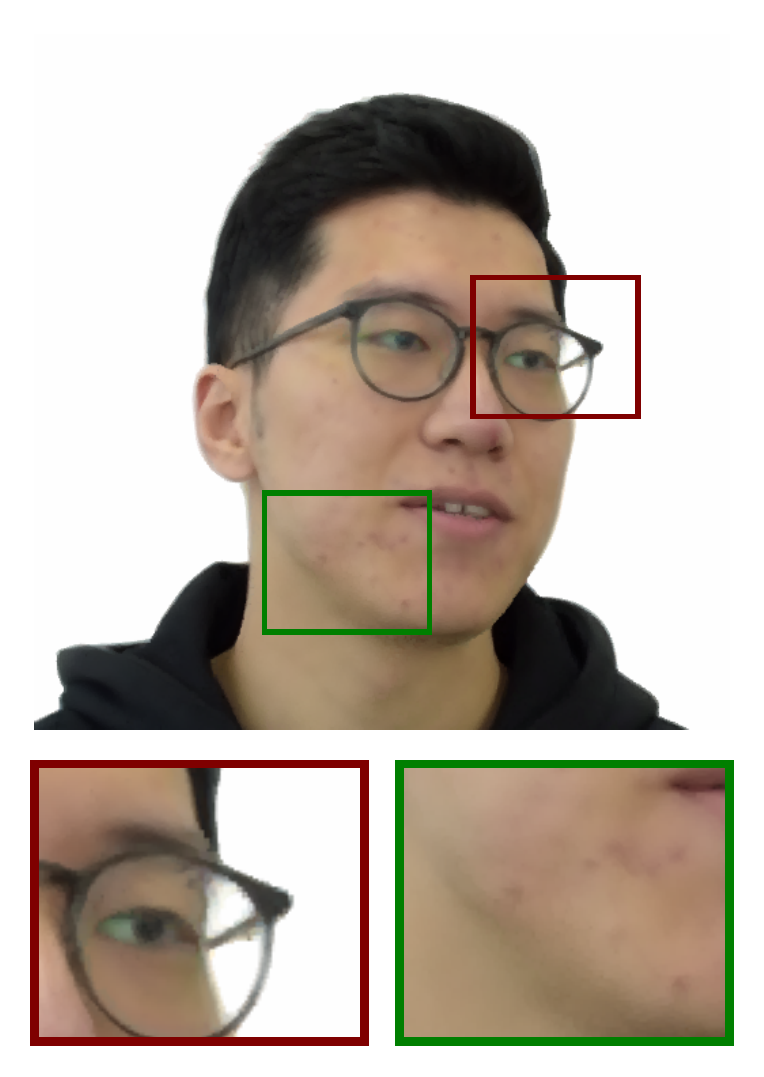}
&\includegraphics[trim=0em 0em 0em 1em, clip=true, height=\myheight]{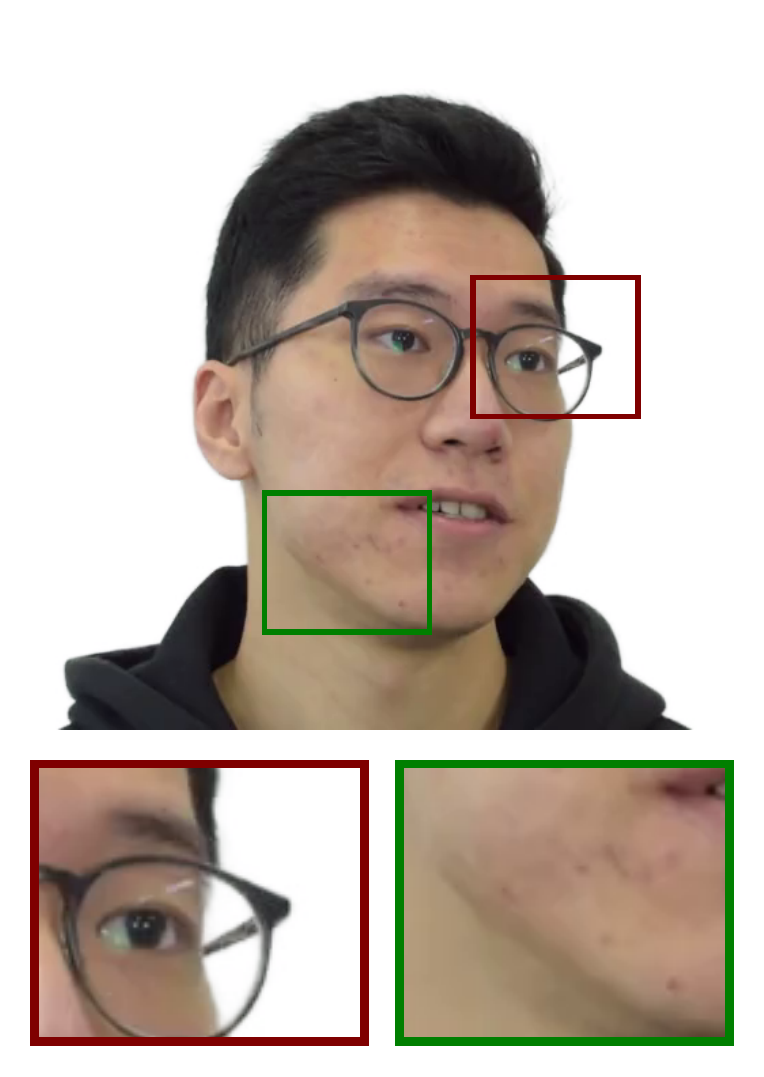}
\\
\includegraphics[height=\myheight]{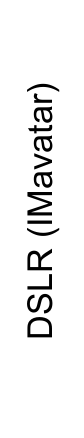}
&\includegraphics[trim=0em 0em 0em 1em, clip=true, height=\myheight]{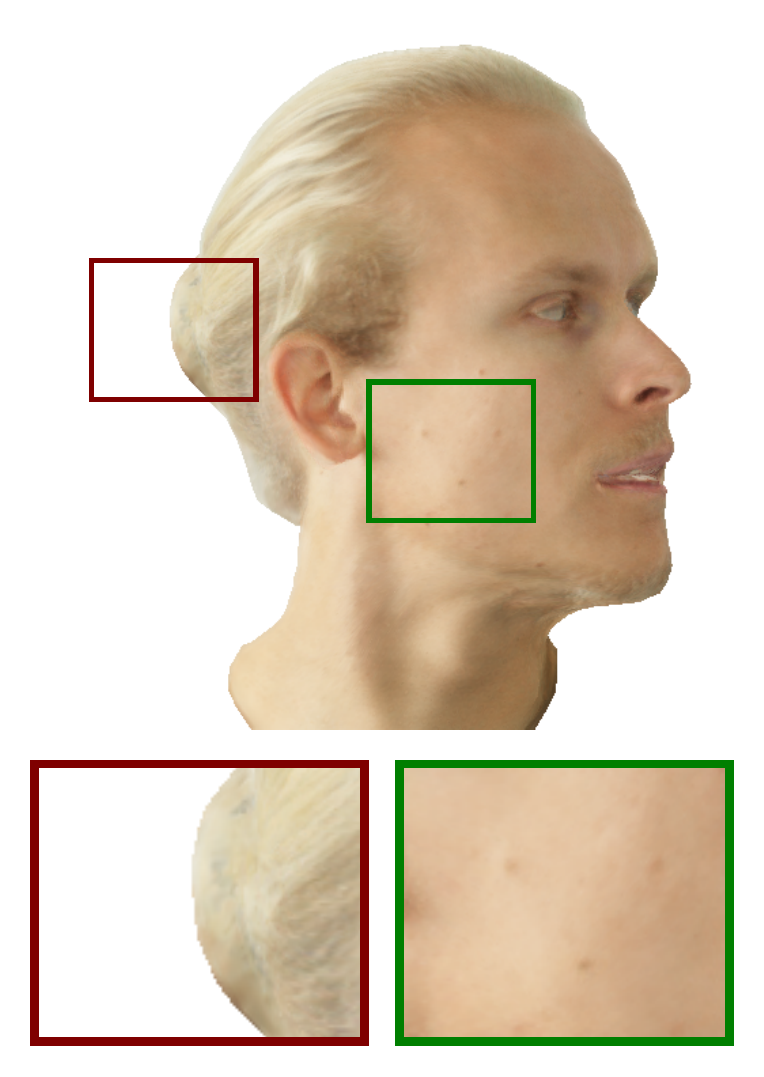}
&\includegraphics[trim=0em 0em 0em 1em, clip=true, height=\myheight]{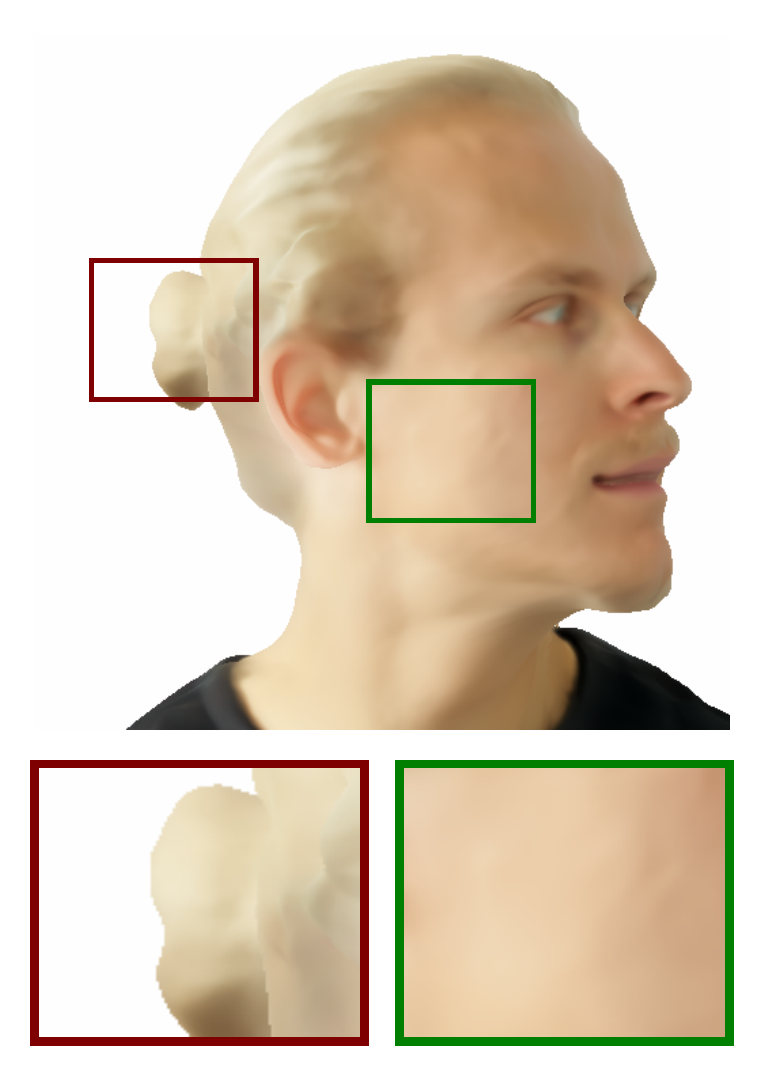}
&\includegraphics[trim=0em 0em 0em 1em, clip=true, height=\myheight]{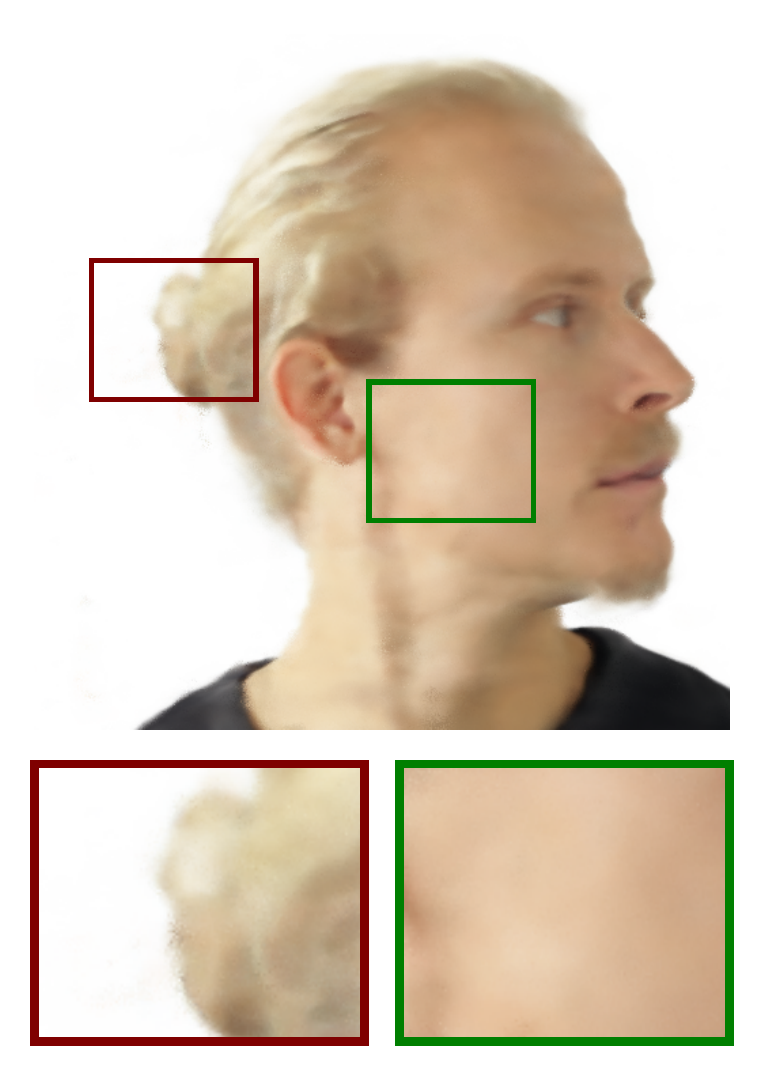}
&\includegraphics[trim=0em 0em 0em 1em, clip=true, height=\myheight]{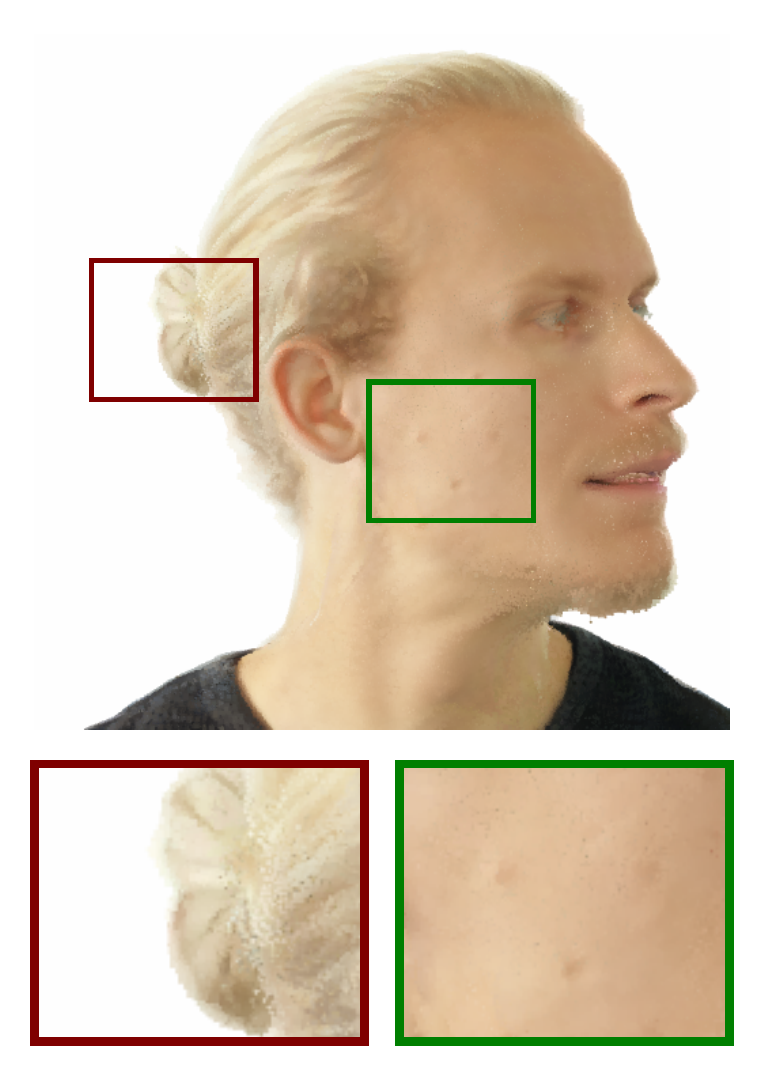}
&\includegraphics[trim=0em 0em 0em 1em, clip=true, height=\myheight]{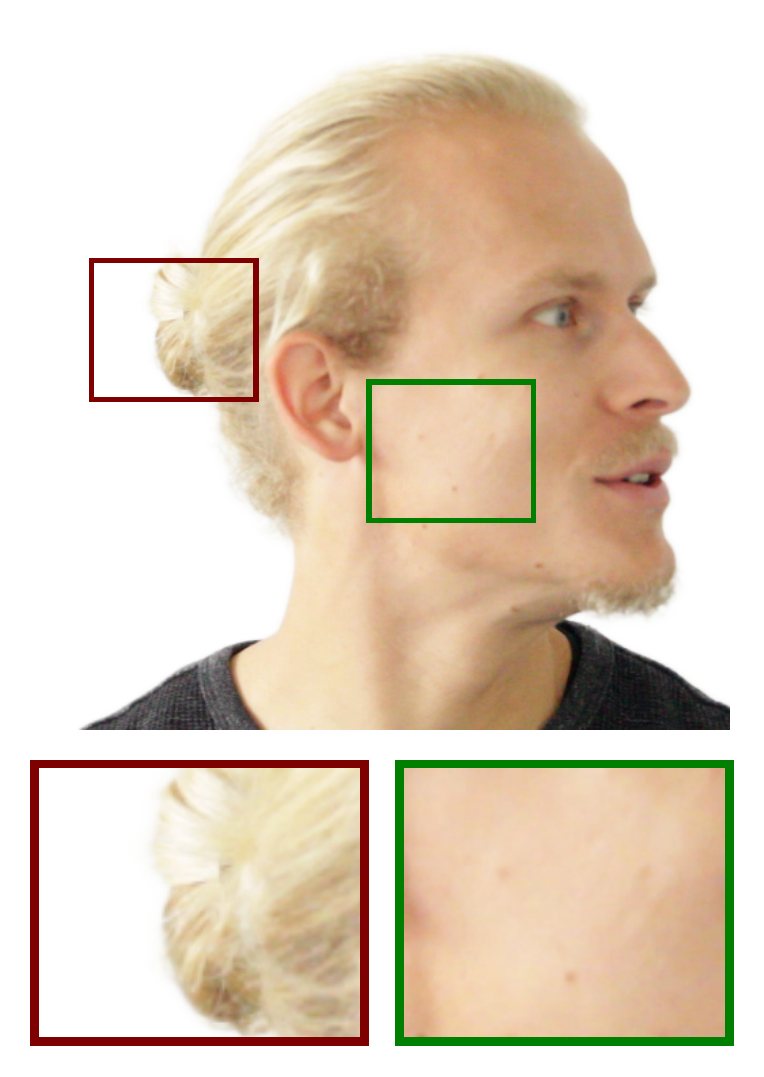}
\\
\includegraphics[height=\myheight]{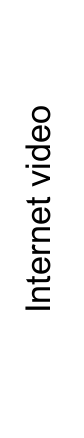}
&\includegraphics[trim=0em 0em 0em 1em, clip=true, height=\myheight]{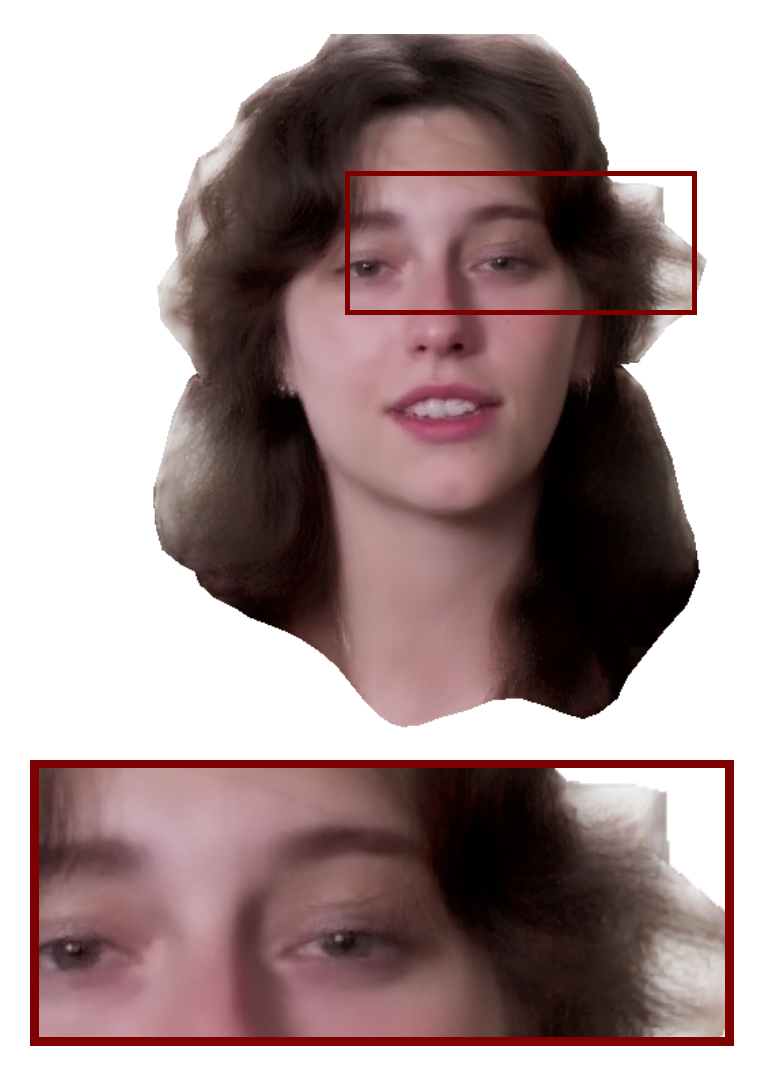}
&\includegraphics[trim=0em 0em 0em 1em, clip=true, height=\myheight]{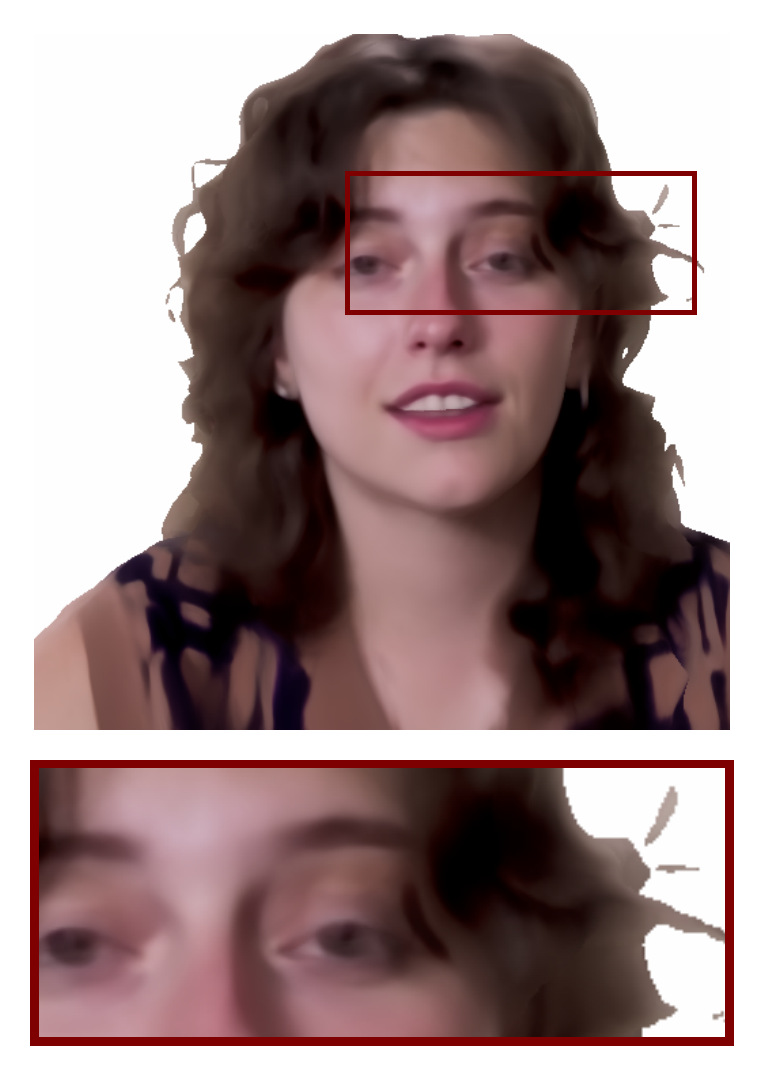}
&\includegraphics[trim=0em 0em 0em 1em, clip=true, height=\myheight]{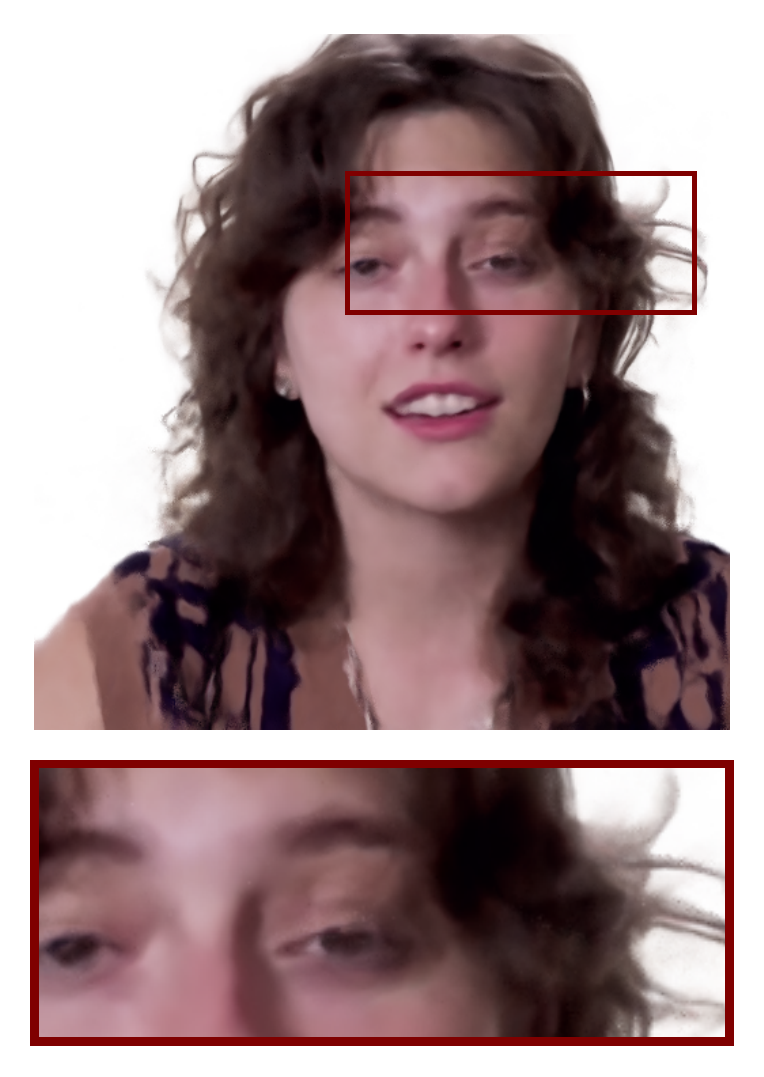}
&\includegraphics[trim=0em 0em 0em 1em, clip=true, height=\myheight]{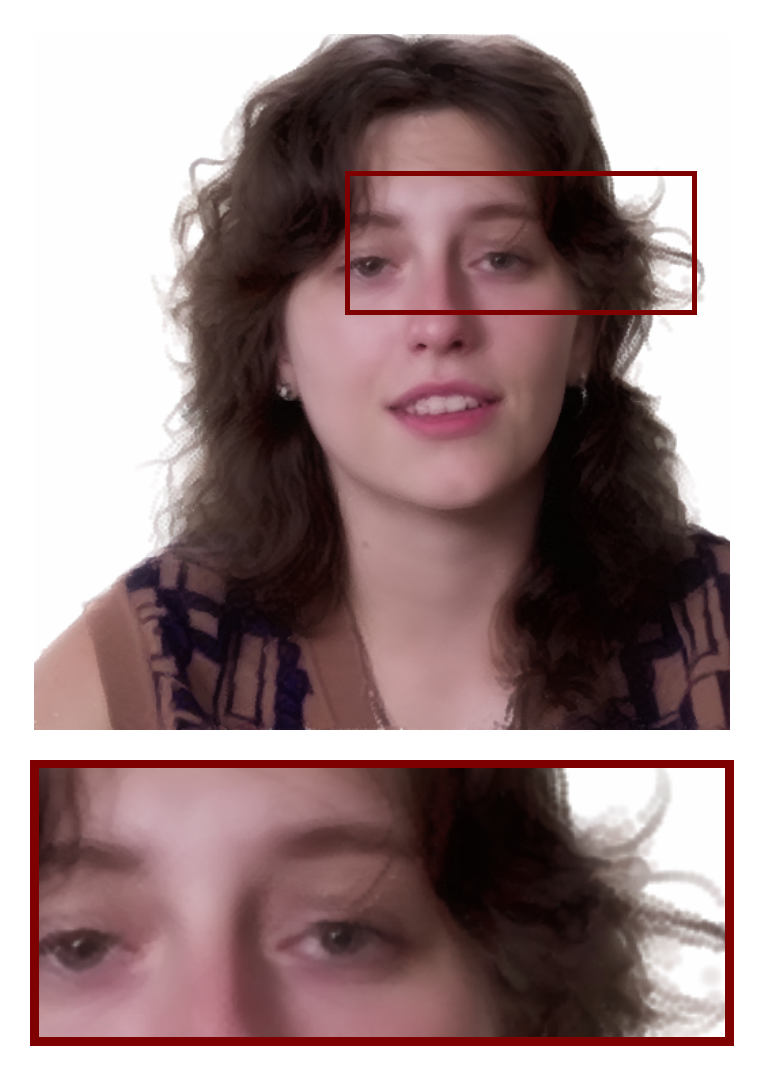}
&\includegraphics[trim=0em 0em 0em 1em, clip=true, height=\myheight]{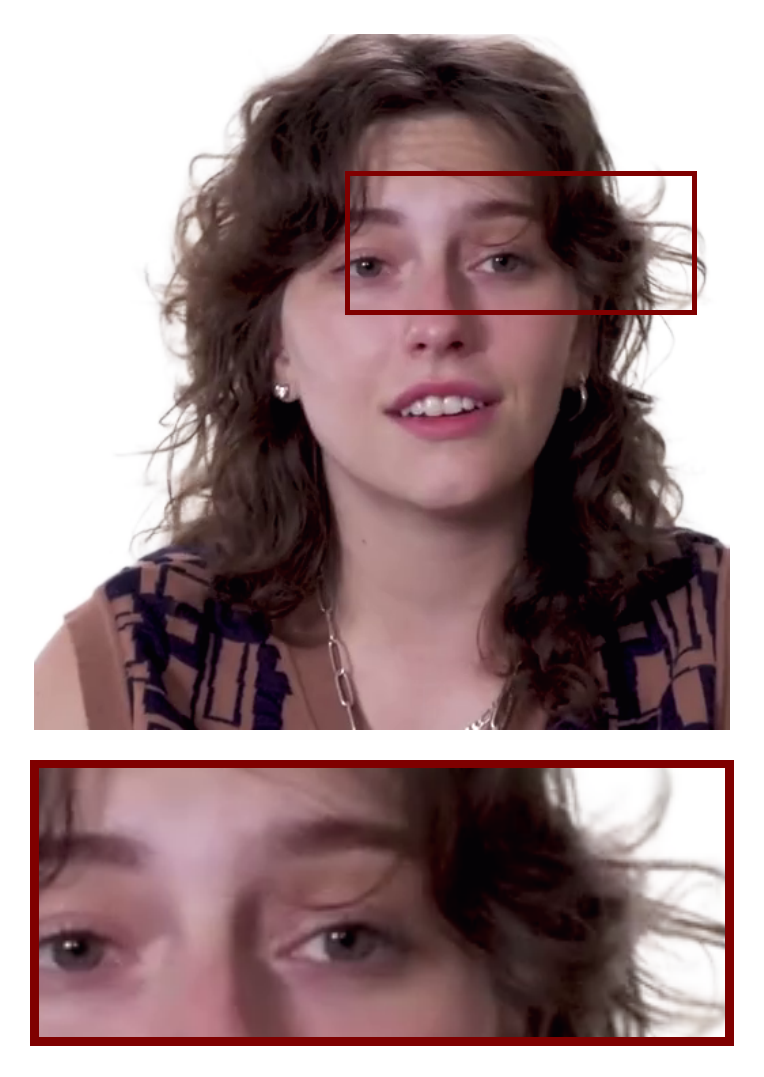}
\\
\includegraphics[height=\myheight]{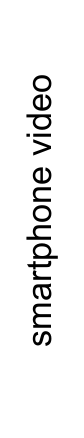}
&\includegraphics[trim=0em 0em 0em 4em, clip=true, height=\myheight]{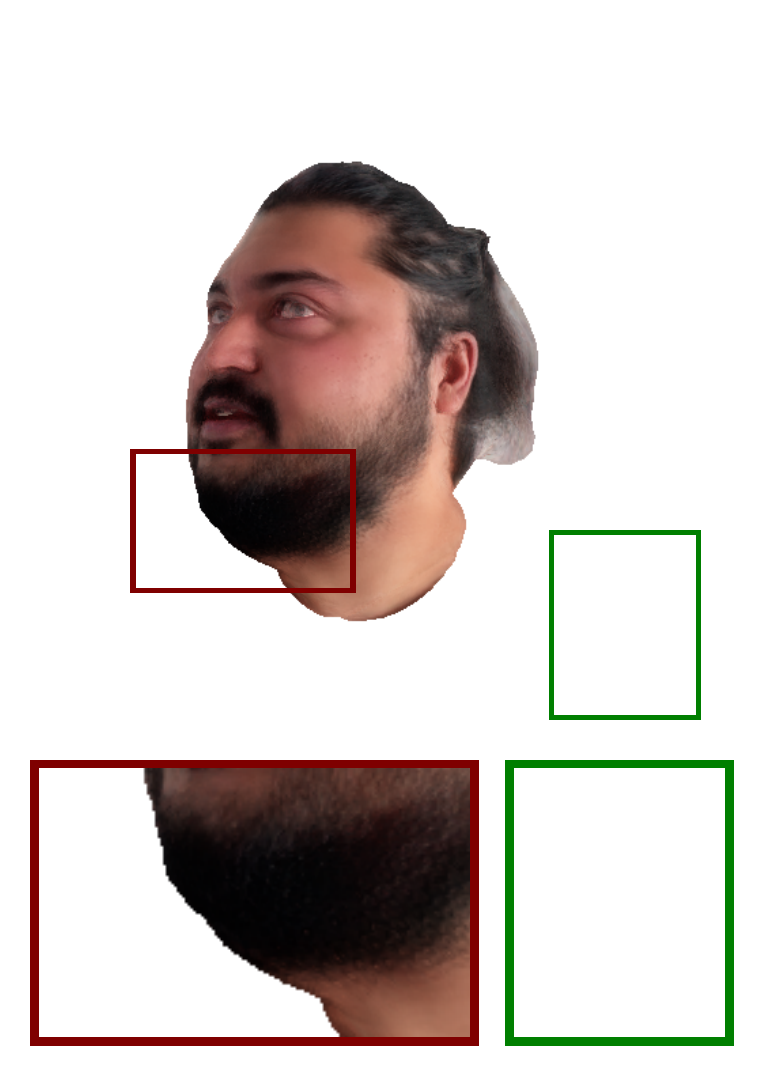}
&\includegraphics[trim=0em 0em 0em 4em, clip=true, height=\myheight]{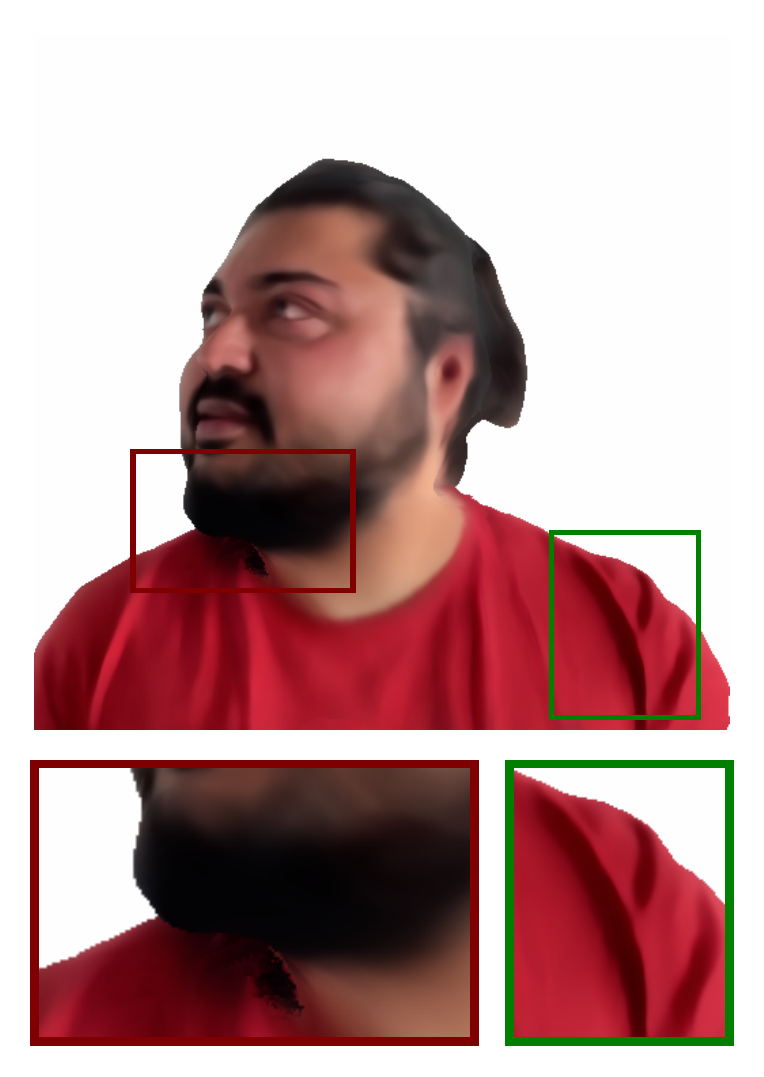}
&\includegraphics[trim=0em 0em 0em 4em, clip=true, height=\myheight]{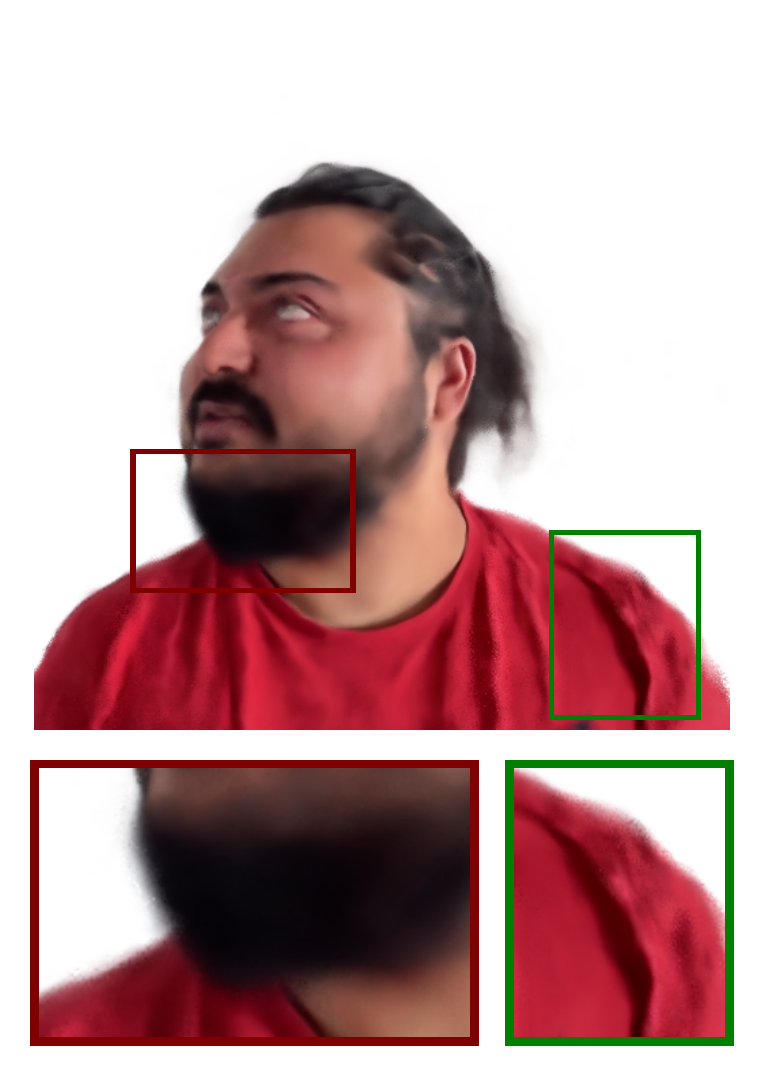}
&\includegraphics[trim=0em 0em 0em 4em, clip=true, height=\myheight]{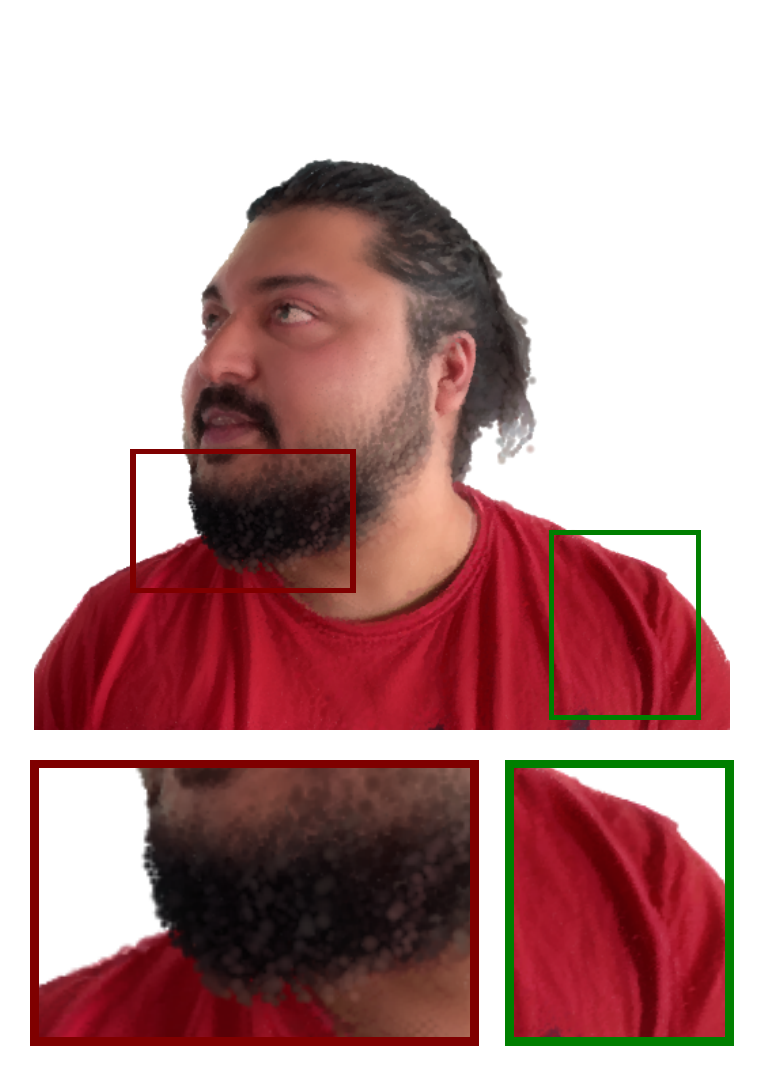}
&\includegraphics[trim=0em 0em 0em 4em, clip=true, height=\myheight]{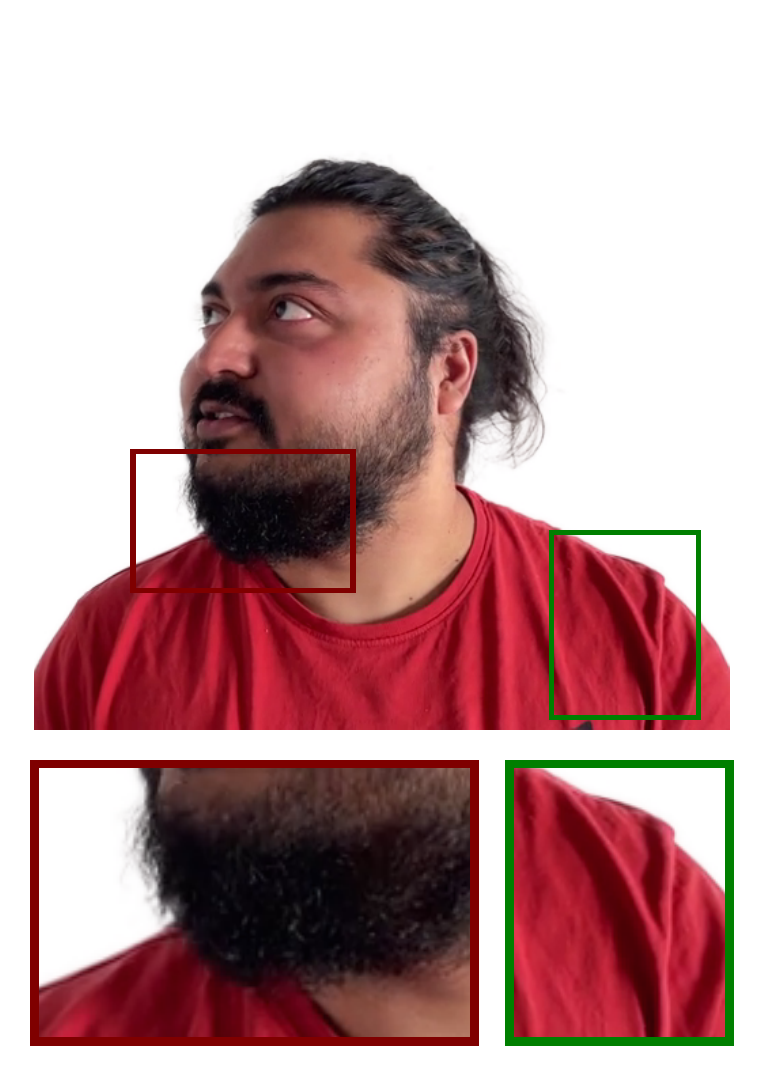}
\\
\includegraphics[height=\myheight]{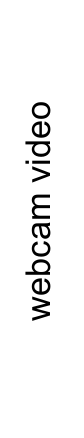}
&\includegraphics[trim=0em 0em 0em 2em, clip=true, height=\myheight]{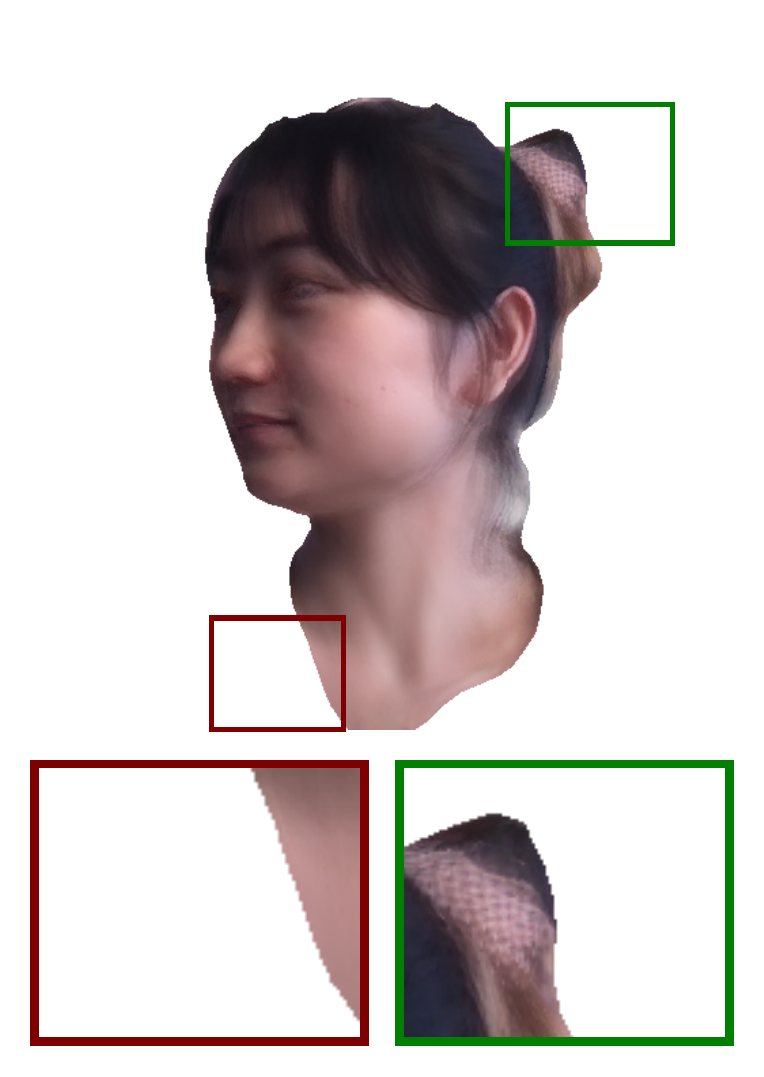}
&\includegraphics[trim=0em 0em 0em 2em, clip=true, height=\myheight]{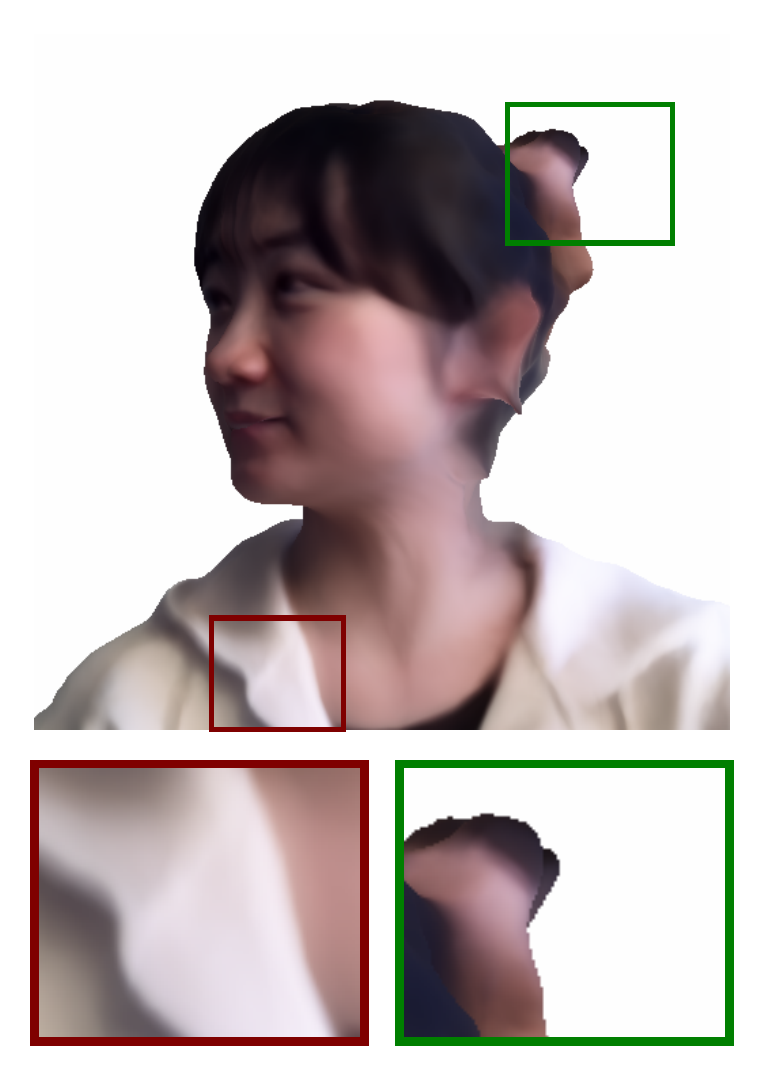}
&\includegraphics[trim=0em 0em 0em 2em, clip=true, height=\myheight]{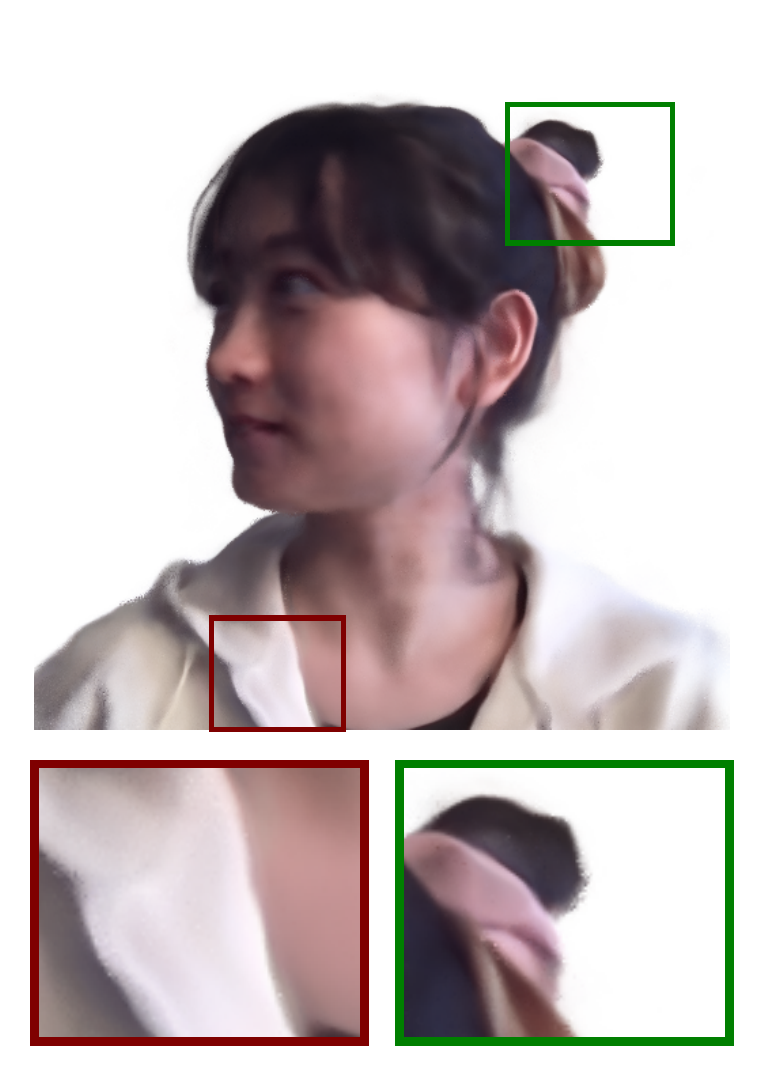}
&\includegraphics[trim=0em 0em 0em 2em, clip=true, height=\myheight]{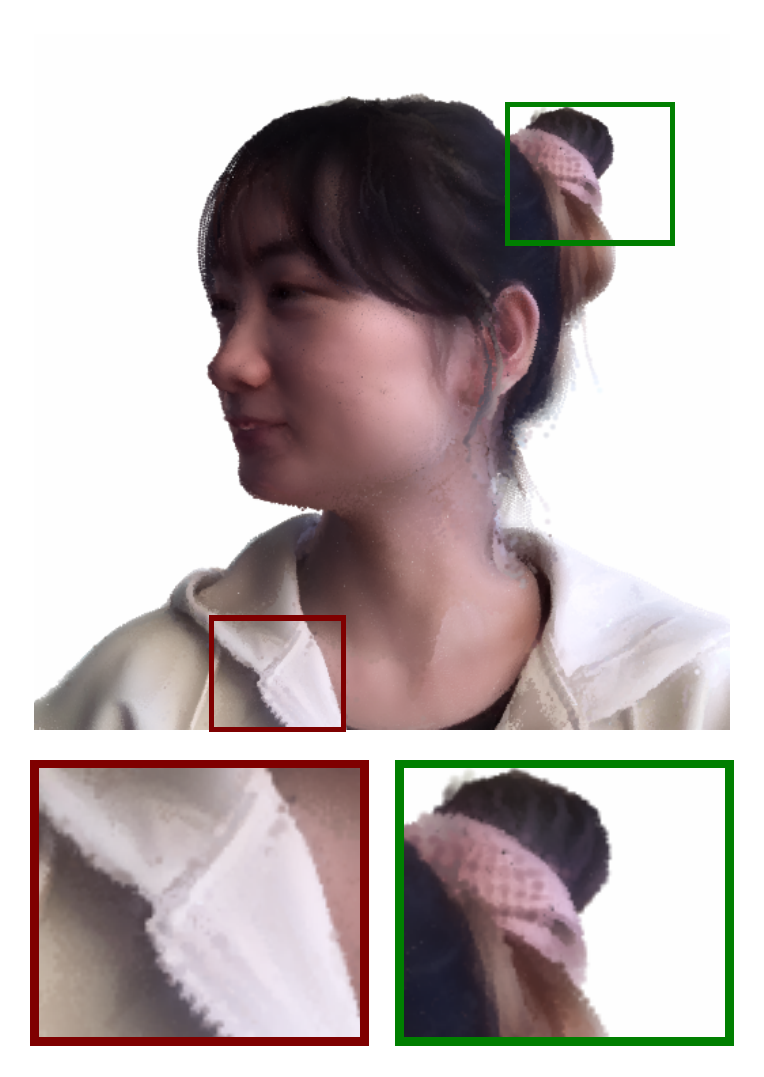}
&\includegraphics[trim=0em 0em 0em 2em, clip=true, height=\myheight]{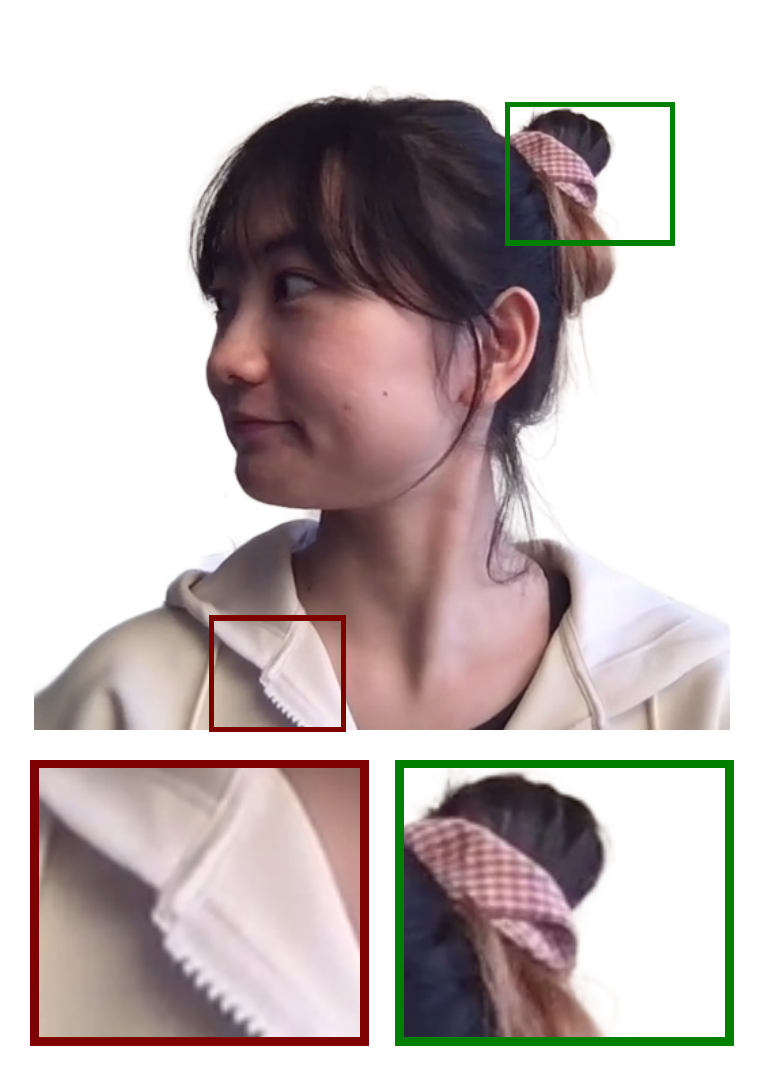}
\\
&NHA~\cite{Grassal_2022_nha} & IMavatar~\cite{Zheng_2022_imavatar} & NerFace~\cite{Gafni_2021_nerface} & \textbf{Ours} & GT
\end{tabularx}
\caption{
\textbf{Qualitative comparison.} \moniker{} produces photo-realistic and detailed appearance compared to SOTA methods, especially apparent in skin details and hair textures. Our point-based method is also flexible enough to capture challenging geometries such as eyeglasses and thin hair strands, which cannot be handled by mesh-based methods.
\label{fig:real_qualitative}
}
\end{center}
\end{figure*}
In Tab.~\ref{tab:sota_real}, we quantitatively compare our point-based avatar method with SOTA baselines using conventional metrics including L1, LPIPS~\cite{zhang18lpips}, SSIM and PSNR. Since NHA~\cite{Grassal_2022_nha} only models the head region, we compare with NHA without considering the clothing region. PointAvatar achieves the best metrics among all methods on both lab-captured DSLR sequences and videos from more casual capture settings, \eg, with smartphones.

\myparagraph{Eyeglasses and detailed structures. }In \cref{fig:real_qualitative}, we show qualitatively that our method can successfully handle the non-face topology introduced by eyeglasses, which pose severe challenges for 3DMM-based methods. 
NHA cannot model the empty space between the frame of the eyeglasses because of its fixed topology and, instead, learns a convex hull painted with dynamic textures. 
In the second example of a man with a hair bun, even though the bun is topologically consistent with the template mesh,
NHA still produces a worse hair silhouette than other methods. 
IMavatar, based on implicit surfaces, is theoretically capable of modeling the geometry of eyeglasses, but it also fails to learn part of the thin frame of glasses. Furthermore, for IMavatar to learn such thin structures, accurate foreground segmentation masks are required, which are hard to obtain in-the-wild. In \suppmat, we show that our method can be trained without mask supervision or known background.

\myparagraph{Surface-like skin and volumetric hair. }Both NHA and IMavatar enforce a surface constraint by design. While such constraint helps with surface geometry reconstruction and multi-view consistency, it also causes problems when modeling volumetric structures such as the curly hair in the third example. Our point-based method is free to model volumetric structures where needed, but also encourages surface-like point geometry in the skin and clothing regions via point pruning, which removes unseen inside points. We show that our method renders sharp and realistic images even for extreme head poses. In contrast, NerFace is a flexible volumetric NeRF~\cite{Mildenhall_2020_nerf}-based avatar method. %
While it is capable of modeling thin structures of any topology, its under-constrained nature leads to poor geometry and compromises rendering quality in uncommon poses. Even for near frontal poses shown in the first and third examples, our method still produces sharper skin details than NerFace.
\subsection{Lighting Disentanglement}\label{sec:relighting}
\setlength{\textfloatsep}{1\baselineskip plus 0.2\baselineskip minus 0.5\baselineskip}
\begin{figure}
\begin{center}
\newcommand{\myheight}{1.6cm}
\setlength\tabcolsep{1pt}
\begin{tabularx}{\textwidth}{ccccc}
\includegraphics[height=\myheight]{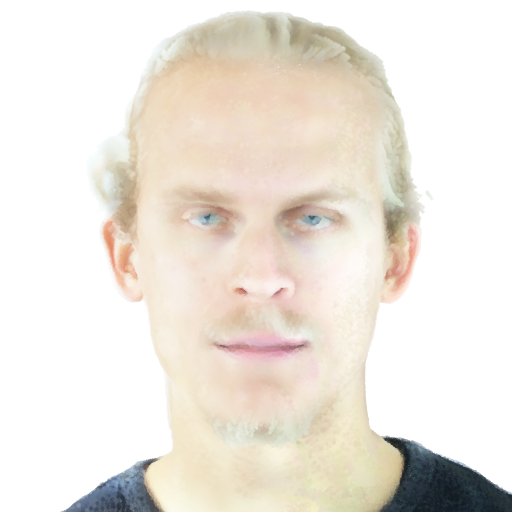}
&\includegraphics[height=\myheight]{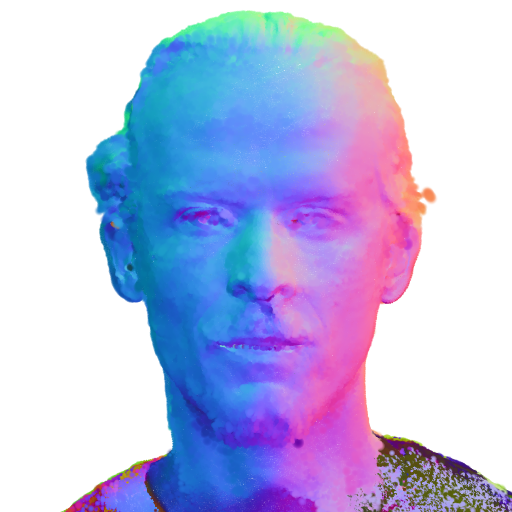}
&\includegraphics[height=\myheight]{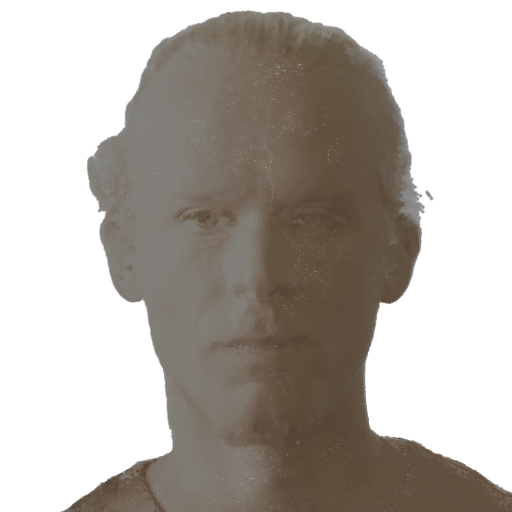}
&\includegraphics[height=\myheight]{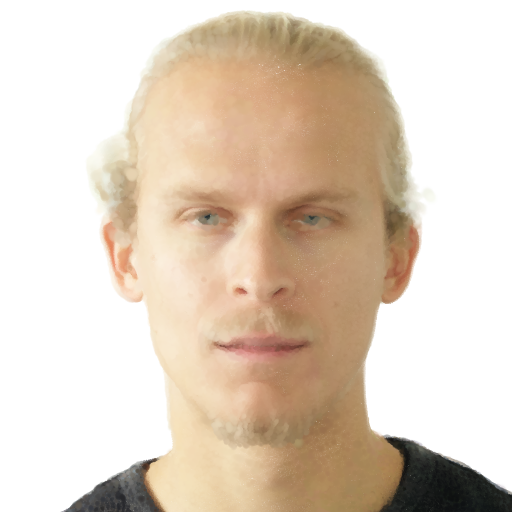}
&\includegraphics[height=\myheight]{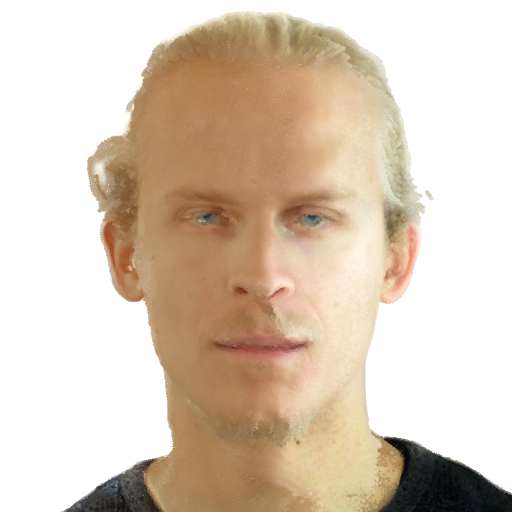}
\\
\includegraphics[height=\myheight]{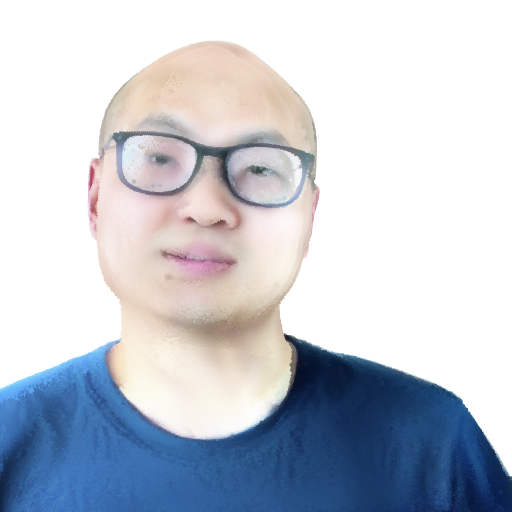}
&\includegraphics[height=\myheight]{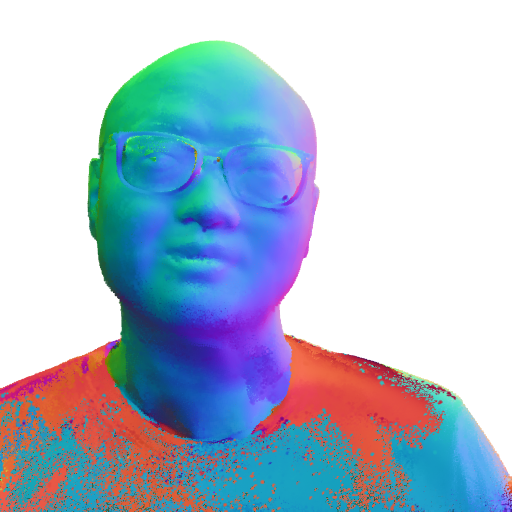}
&\includegraphics[height=\myheight]{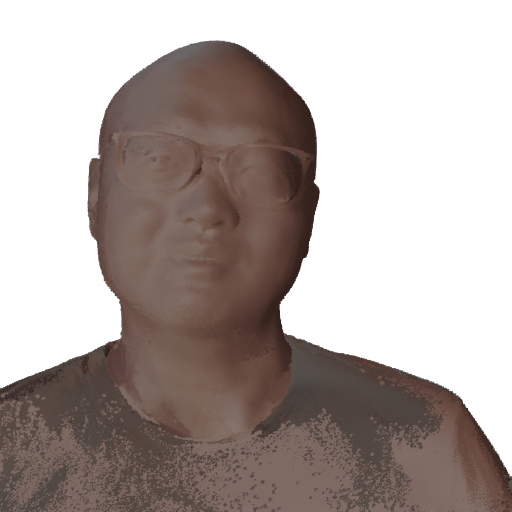}
&\includegraphics[height=\myheight]{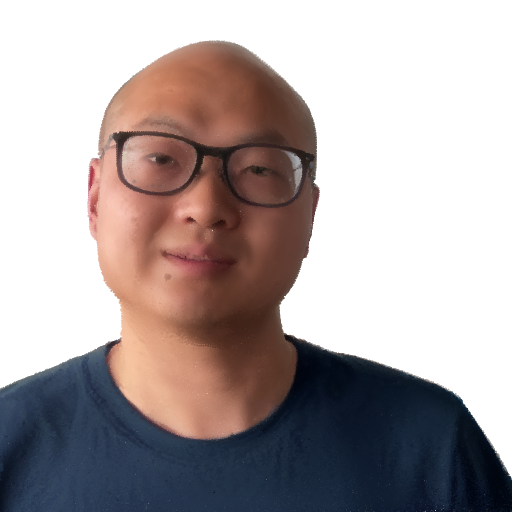}
&\includegraphics[height=\myheight]{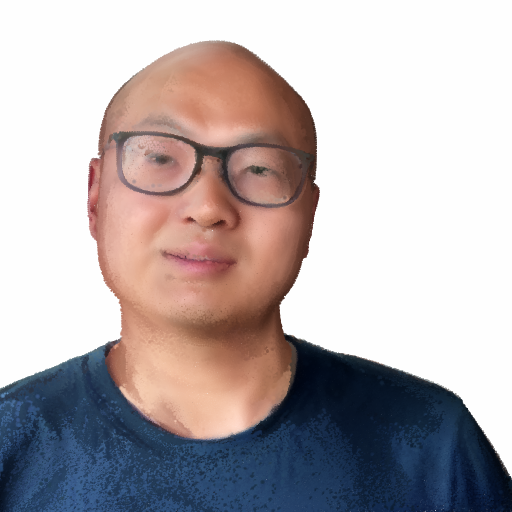}
\\
albedo & normal & shading & rendering & relit
\end{tabularx}
\caption{
\textbf{Self-supervised lighting disentanglement.} \moniker{} disentangles albedo and normal-depedent shading from a single video captured with a fixed lighting condition. After training, \moniker{} can be faithfully relit.
\label{fig:lighting}
}
\end{center}
\end{figure}
Our method disentangles rendered colors into intrinsic albedo colors and normal-dependent shading values. By changing the shading component, we can alter the lighting directions as shown in Fig.~\ref{fig:lighting} (see \suppmat{} for details of relighting). In the following, we demonstrate that lighting disentanglement, along with the proposed normal estimation techniques, improves the geometry details of the reconstruction.
In column 2 of Fig.~\ref{fig:geometry_ablation}, we show that, disentangling shading and albedo itself improves facial geometry (see the cheek area). 
We further ablate two of our design choices for obtaining better point normals. 
First, we compare our SDF-based normal estimation with direct normal estimation, which calculates normals by approximating the local plane of neighboring points. 
Normals obtained through SDF are less noisy, especially in highly detailed regions such as the hair, eyes and mouth, where direct normal estimation often fails. 
Second, we compare our proposed normal transformation using the deformation Jacobian vs. transforming the normals simply with the rotation matrix of the point deformation. The latter ignores the spatial changes of blendshapes and LBS weights.
As column 4 shows, our normal transformation method takes into account the effects of additive blendshapes, and is able to produce correct normals around the nasal line when smiling. 
\begin{figure}
\vskip -3mm
\begin{center}
\newcommand{\myheight}{2.2cm}
\setlength\tabcolsep{2pt}
\begin{tabularx}{\textwidth}{cccc}
\includegraphics[trim=4em 0em 10em 0em, clip=true, height=\myheight]{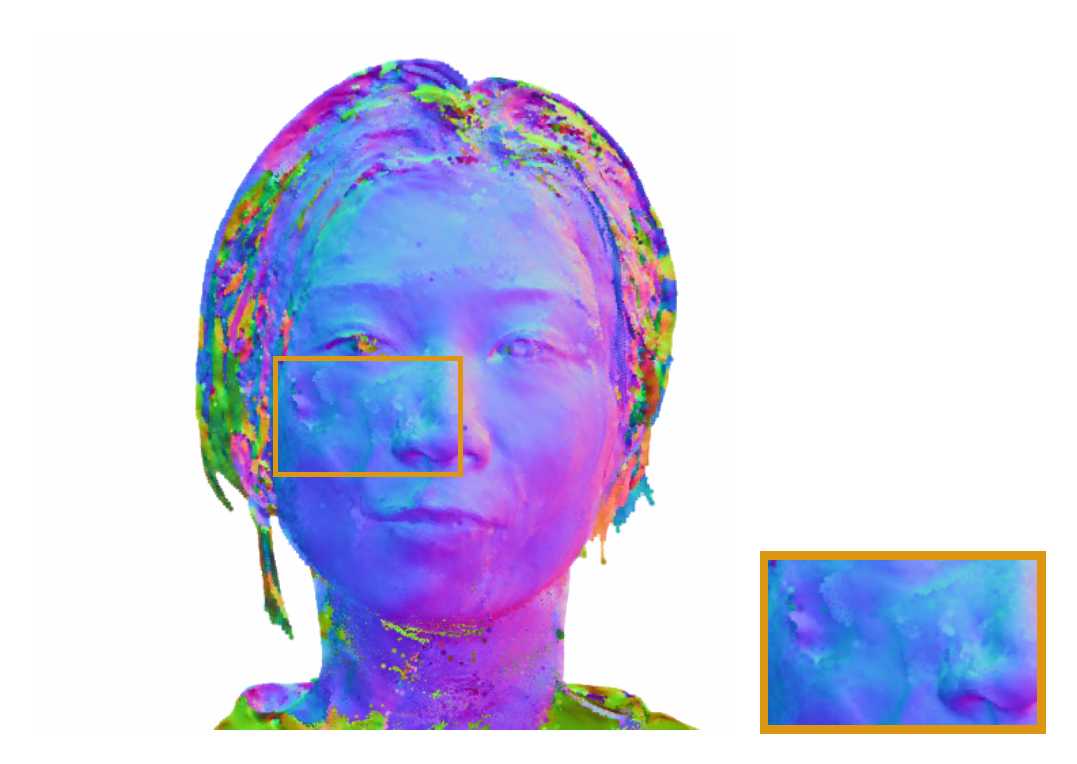}
&\includegraphics[trim=4em 0em 10em 0em, clip=true, height=\myheight]{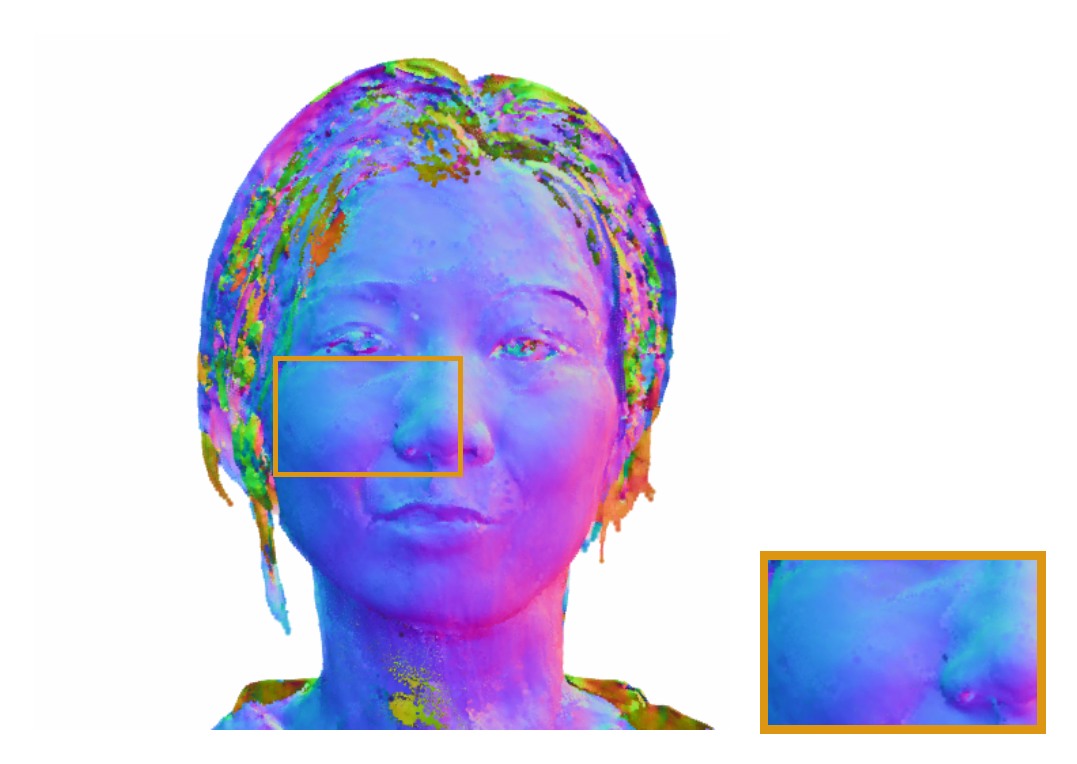}
&\includegraphics[trim=4em 0em 10em 0em, clip=true, height=\myheight]{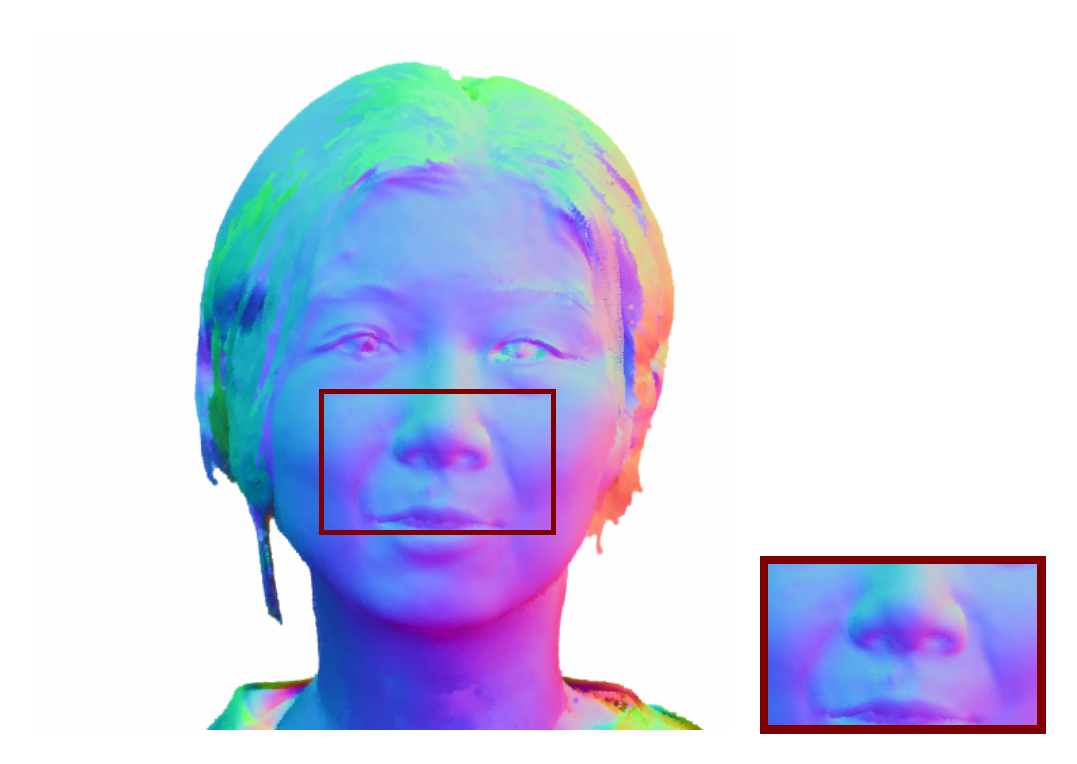}
&\includegraphics[trim=4em 0em 10em 0em, clip=true, height=\myheight]{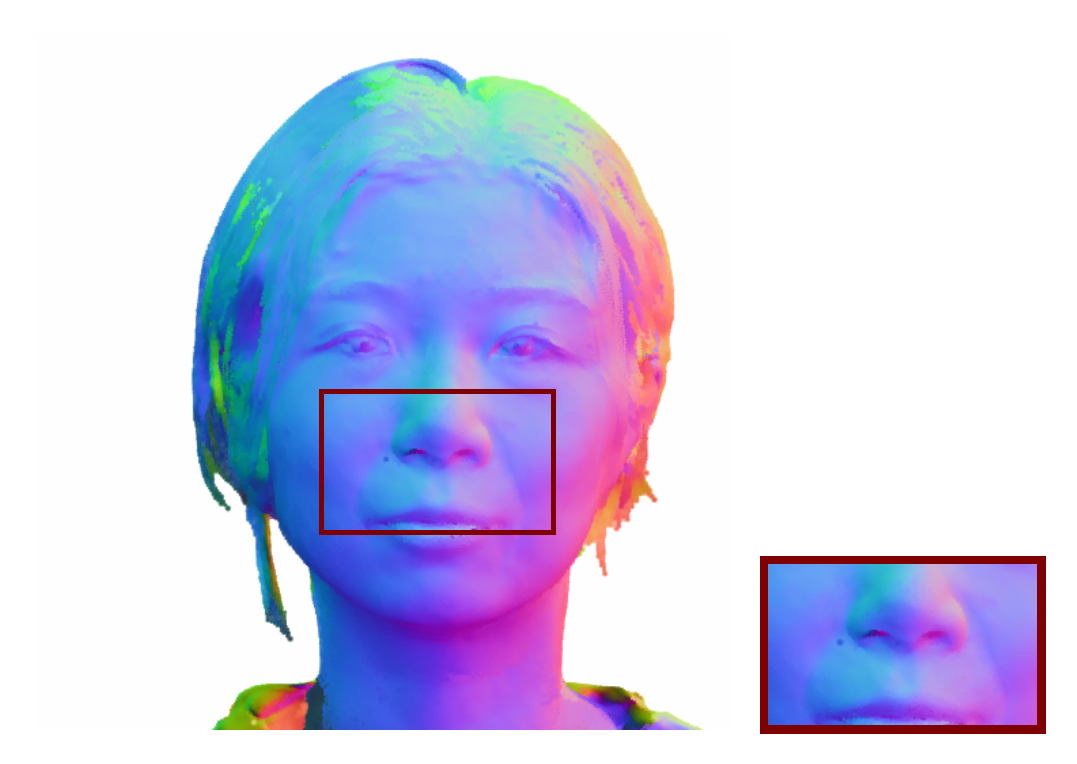}
\\
\includegraphics[trim=4em 0em 10em 0em, clip=true, height=\myheight]{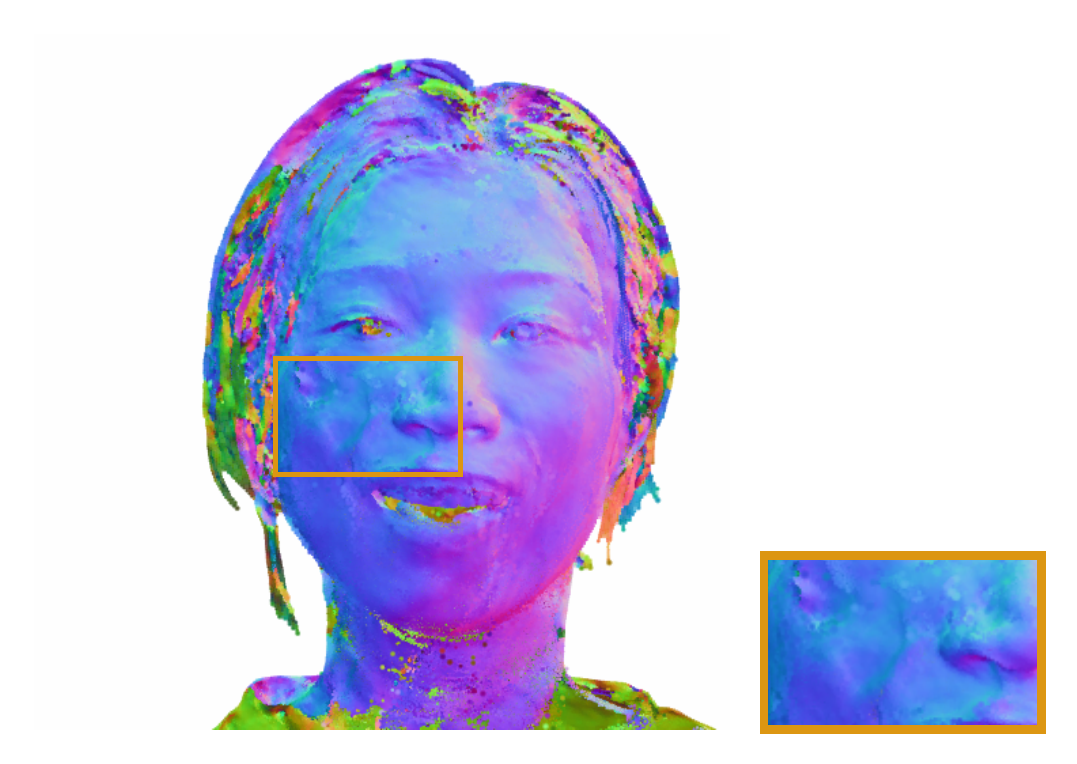}
&\includegraphics[trim=4em 0em 10em 0em, clip=true, height=\myheight]{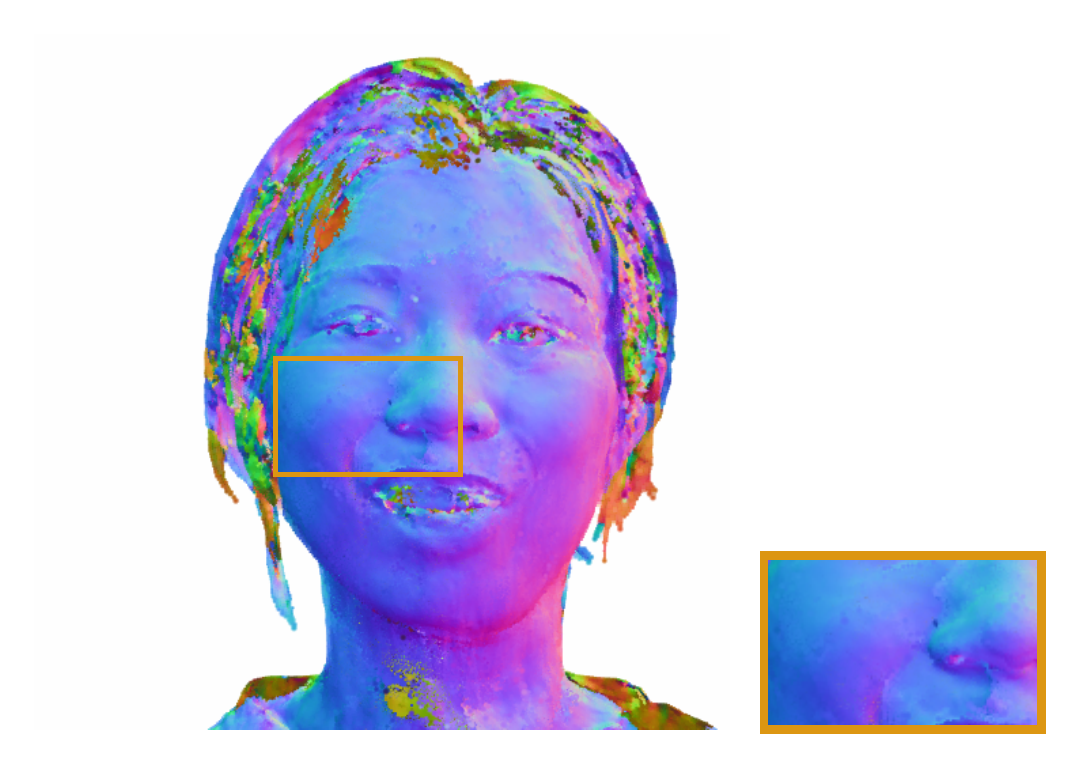}
&\includegraphics[trim=4em 0em 10em 0em, clip=true, height=\myheight]{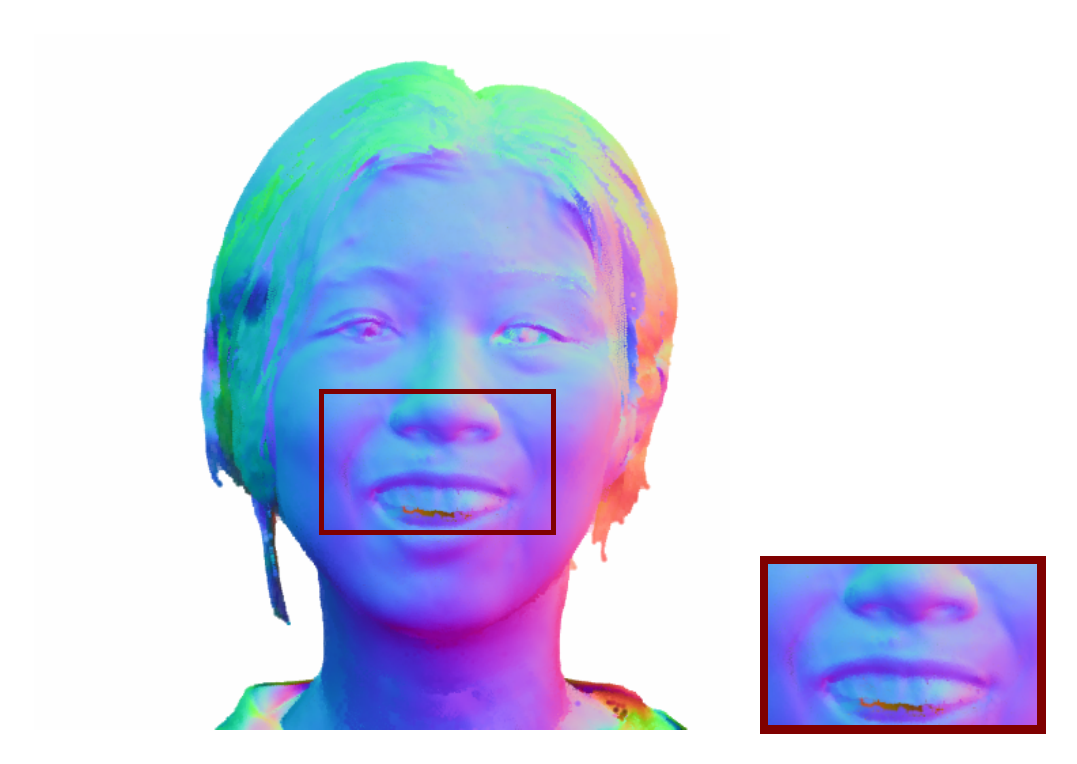}
&\includegraphics[trim=4em 0em 10em 0em, clip=true, height=\myheight]{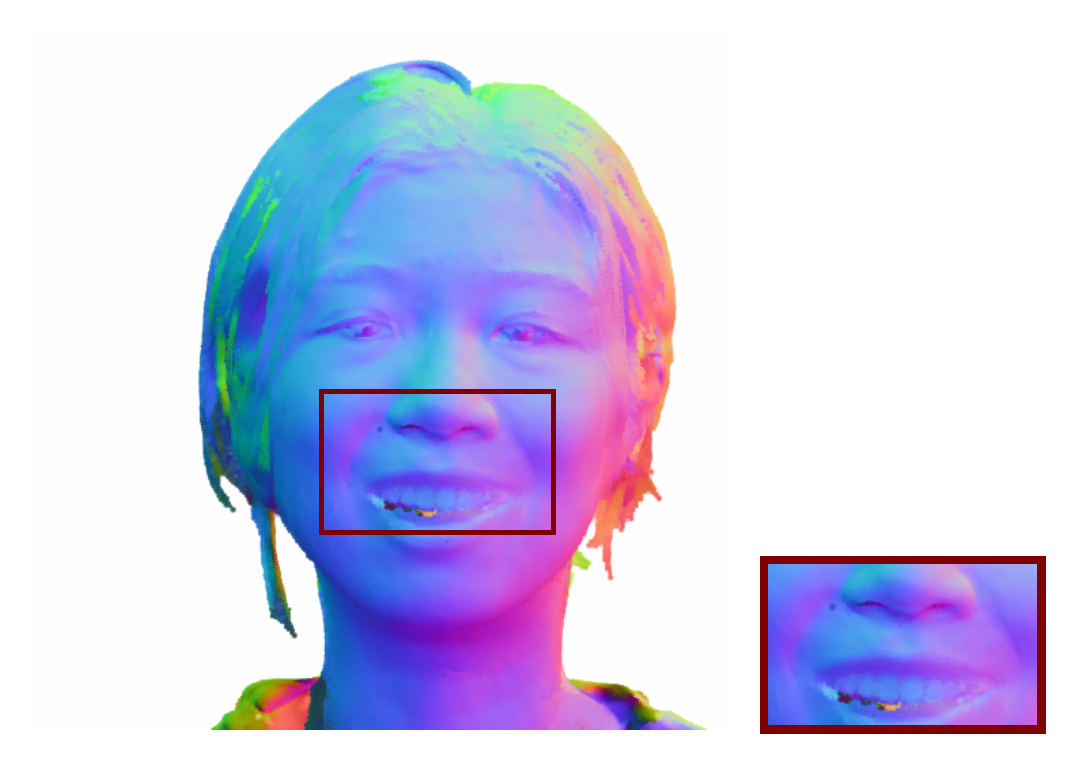}
\\
(1) entangled & (2) 1+disentangle & (3) 2+SDF & (4) 3+Jacobian
\end{tabularx}
\caption{
\textbf{Geometry ablation. } Disentangling shading and albedo improves facial normals compared to an entangled rendering model. Moreover, obtaining canonical point normals from SDF further improved smoothness. Transforming normals with the spatial Jacobian of the deformation field enables the capture of blendshape-related normal changes, \eg, around the nasal line.
\label{fig:geometry_ablation}
}
\end{center}
\end{figure}

\subsection{Training Efficiency}
\begin{table}
\resizebox{\linewidth}{!}{
    \begin{tabular}{lcc}
    \toprule
    Method & Training time (hour) & Rendering time per image (train)\\
    \midrule
    IMavatar & 48h & 100s \\
    NerFace & 54h & 4s \\
    Ours & 6h & 0.1s - 1.5s (varies with point numbers) \\ 
    \bottomrule
    \end{tabular}
}
\caption{\textbf{Training and rendering time.} Compared to implicit-based methods, our method is significantly faster in training and is able to render full-images much more efficiently. }
\label{tab:efficiency}
\end{table}

\begin{figure}
\begin{center}
\newcommand{\myheight}{1.9cm}
\setlength\tabcolsep{1pt}
\begin{tabularx}{\textwidth}{cccc}
\includegraphics[trim=0em 3em 0em 1em, clip=true, height=\myheight]{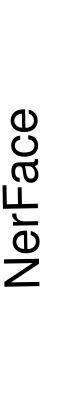}
&\includegraphics[height=\myheight]{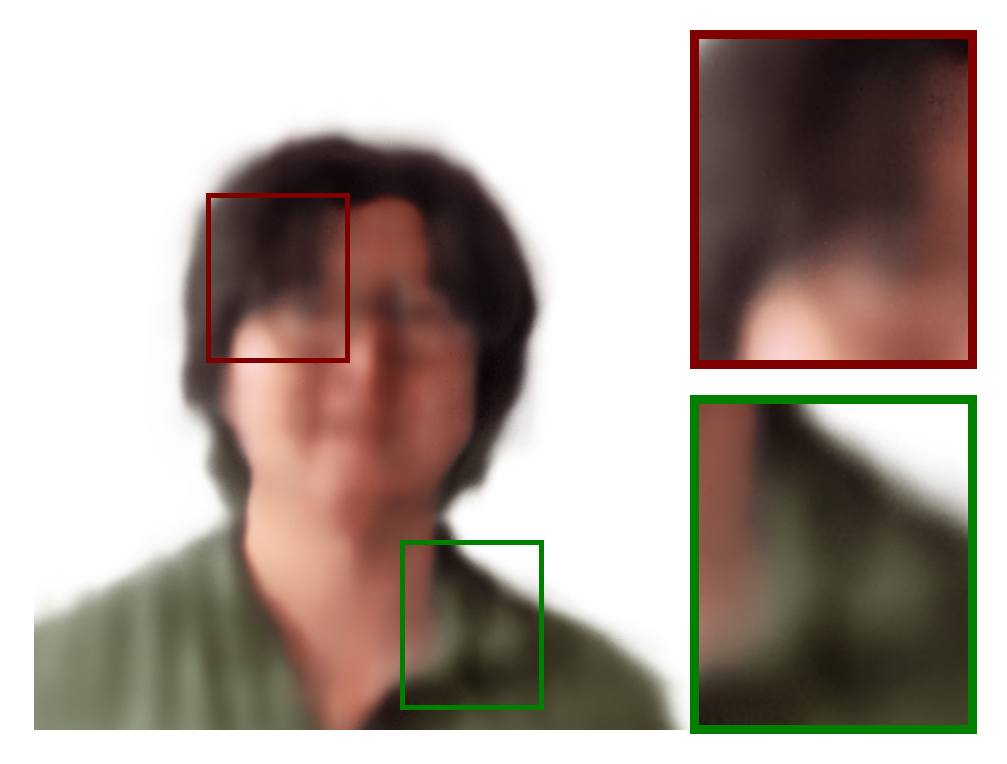}
&\includegraphics[height=\myheight]{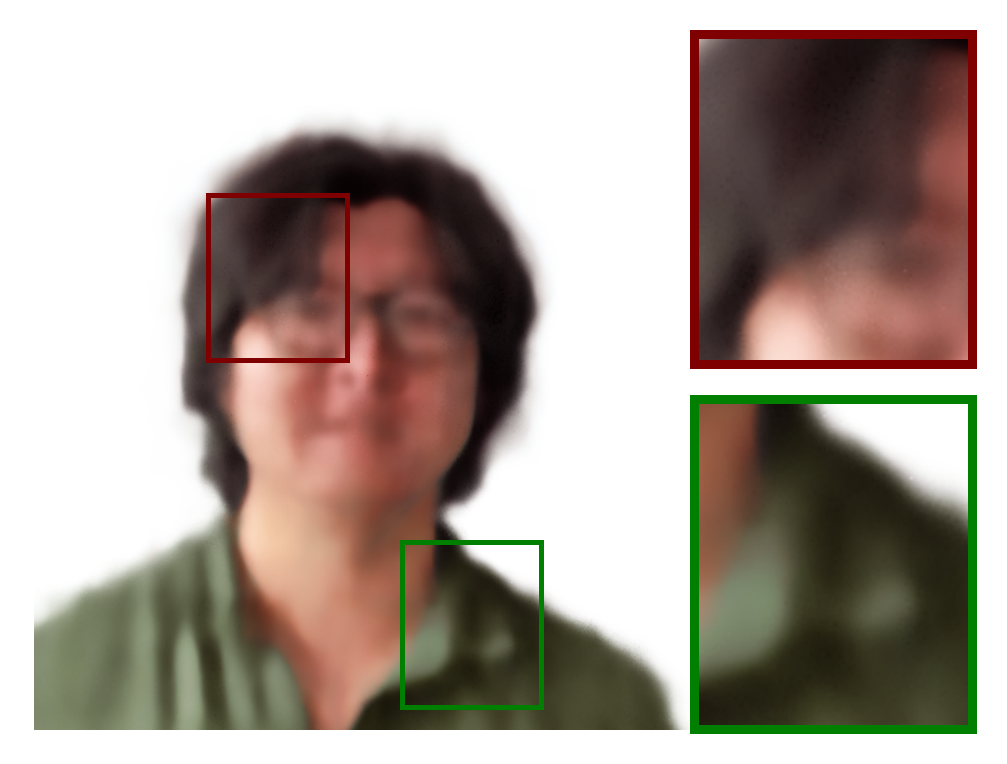}
&\includegraphics[height=\myheight]{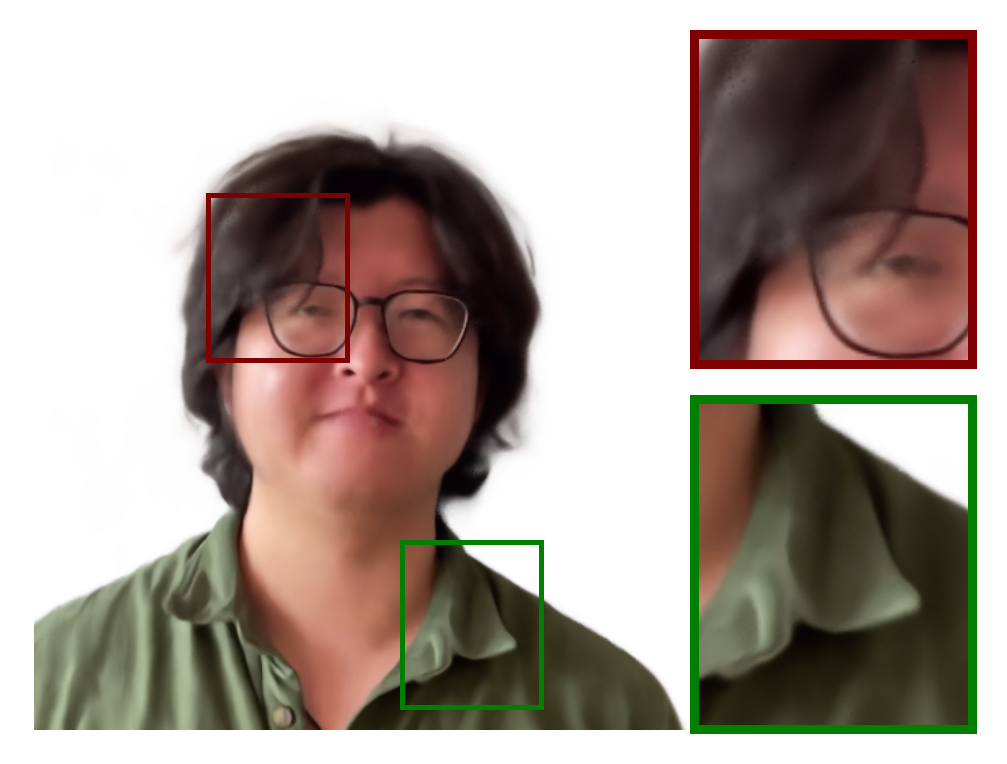}
\\
&\small{20.82}
&\small{20.78}
&\small{21.11}
\\
\includegraphics[trim=0em 3em 0em 1em, clip=true, height=\myheight]{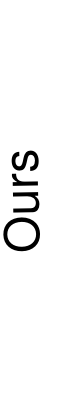}
&\includegraphics[height=\myheight]{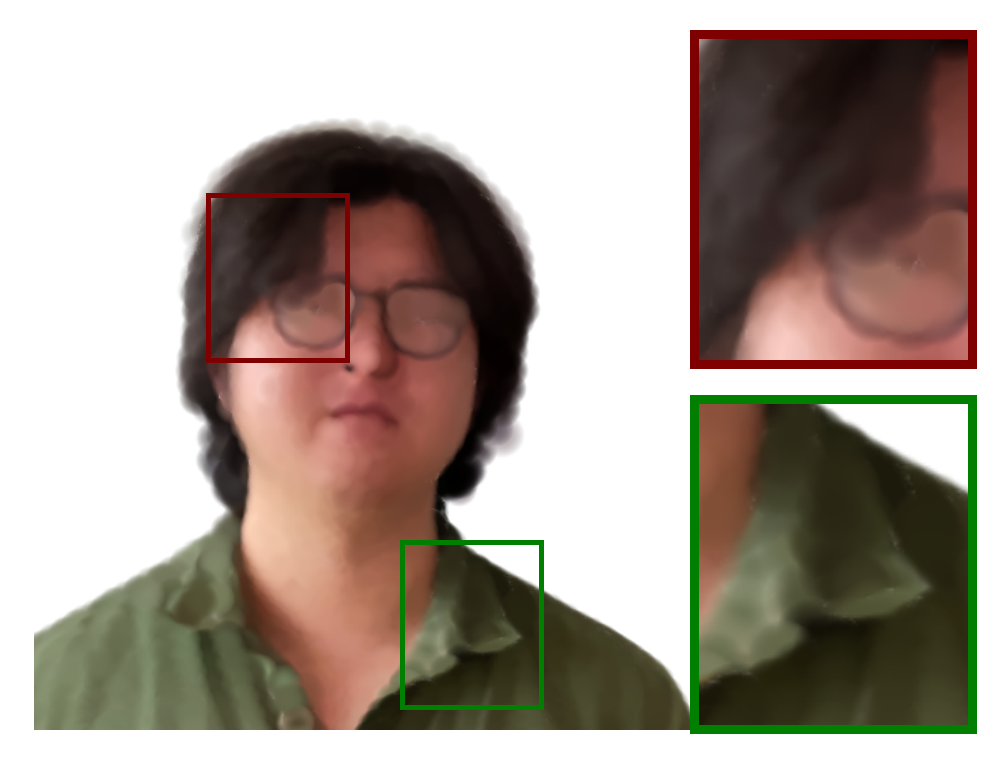}
&\includegraphics[height=\myheight]{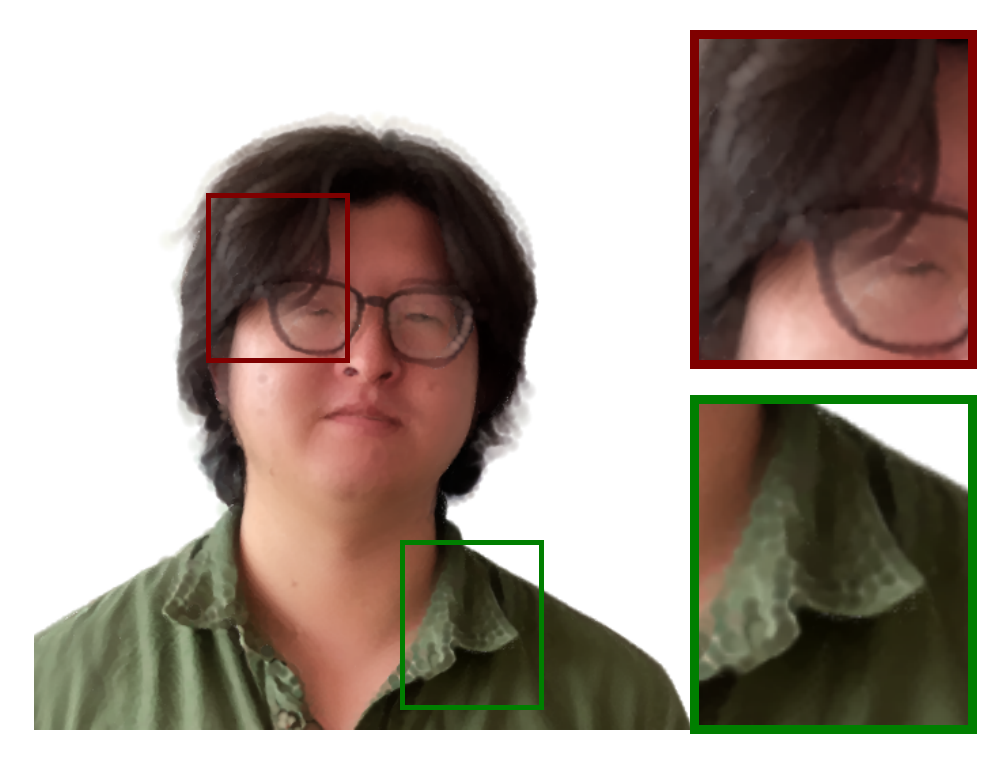}
&\includegraphics[height=\myheight]{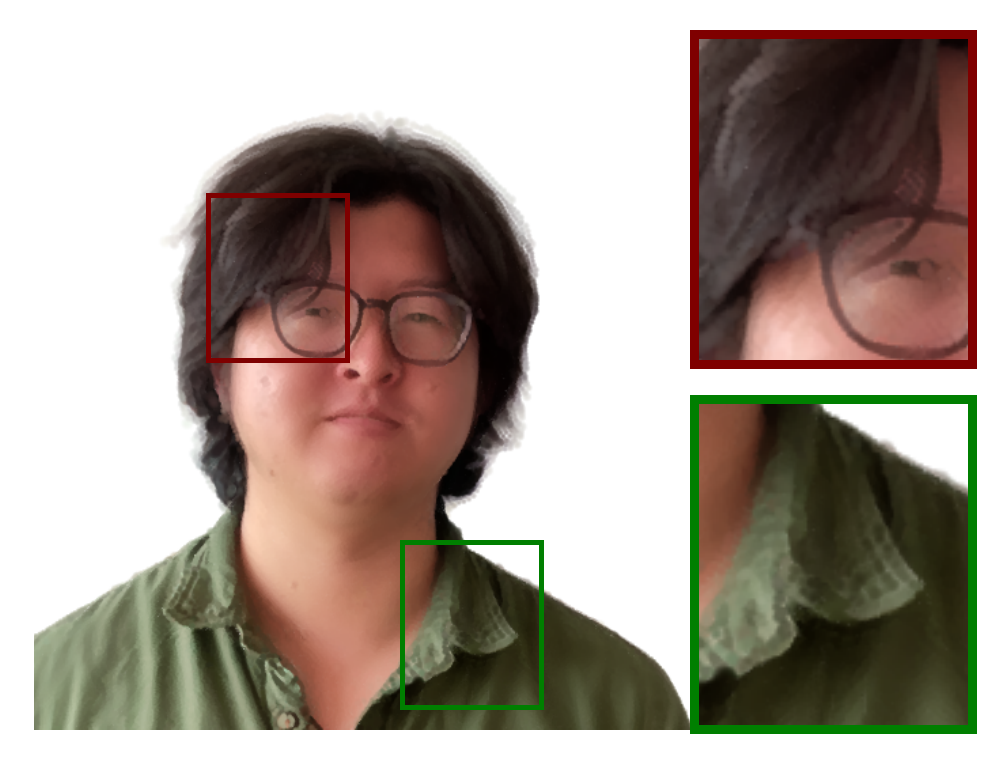}
\\
&\small{23.48}
&\small{24.52}
&\small{24.30}
\\
& 40m & 1h30m & 3h %
\\
\end{tabularx}
\caption{
\textbf{Training efficiency.} \moniker{} converges much faster than implicit-based methods. Here we show NerFace~\cite{Gafni_2021_nerface} as an example, and indicate the average PSNR of the test set.
\label{fig:speed}
}
\end{center}
\end{figure}
\begin{figure}
\begin{center}
\newcommand{\myheight}{2.cm}
\setlength\tabcolsep{1pt}\centering
\begin{tabularx}{\textwidth}{cc}
\includegraphics[trim=-10em 0em -3em 0em, clip=true, height=\myheight]{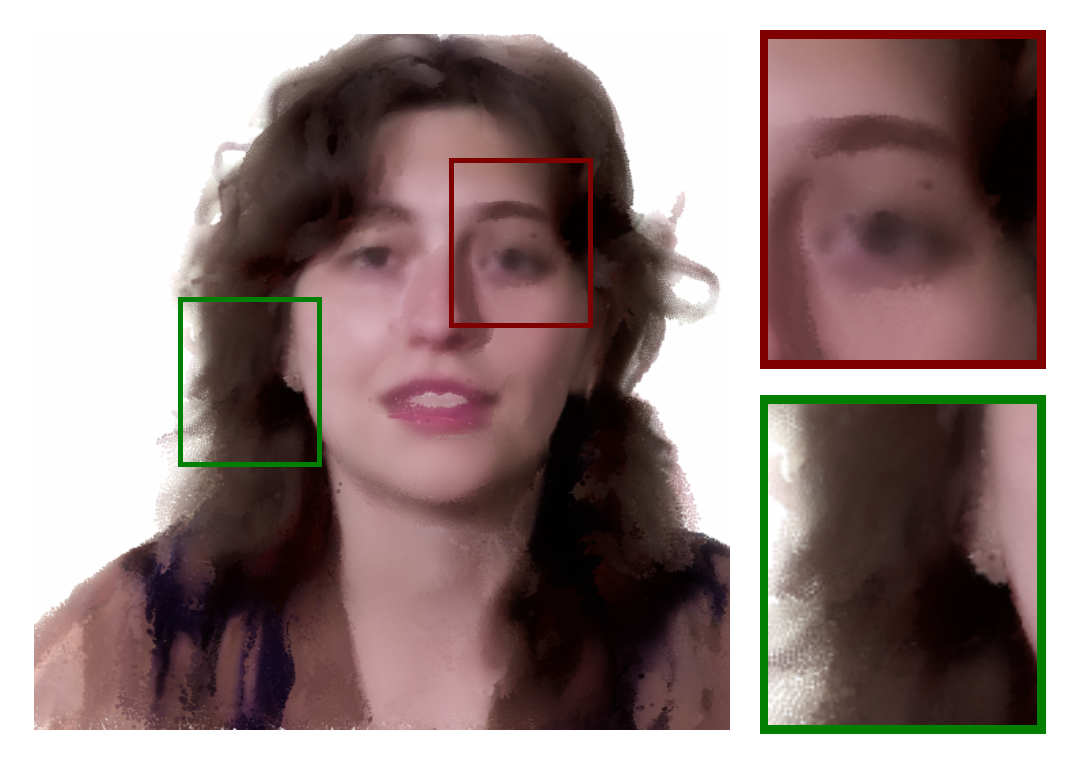}
&\includegraphics[trim=-3em 0em -10em 0em, clip=true, height=\myheight]{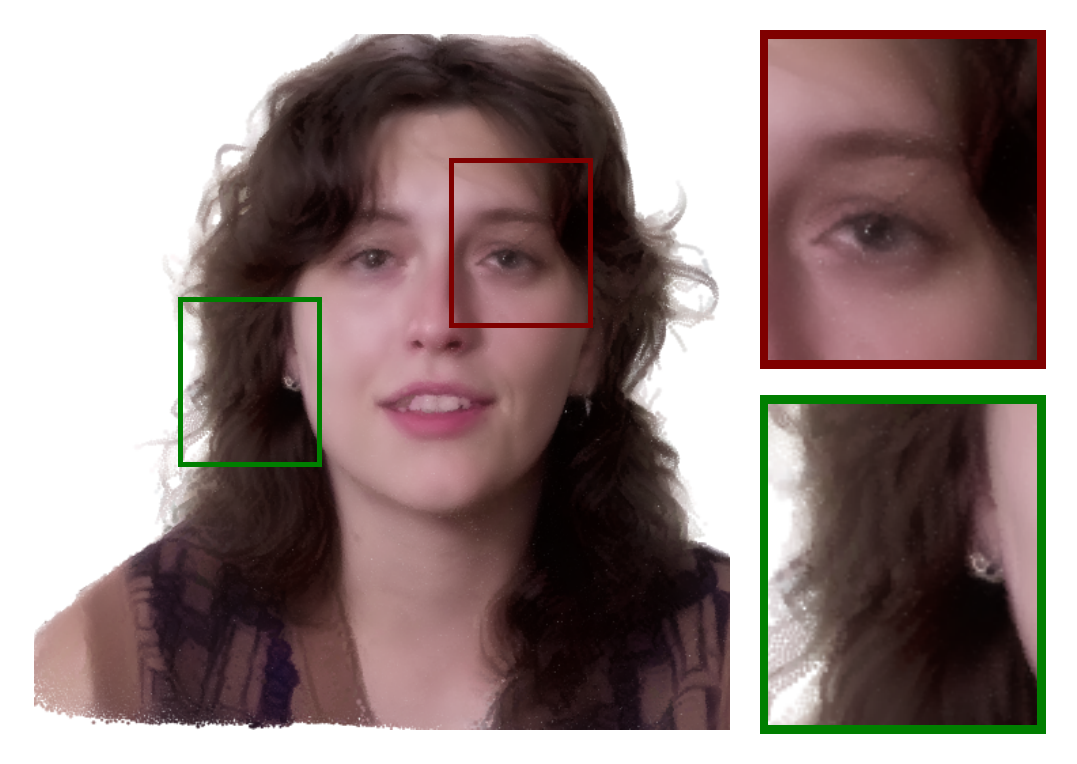}
\\
wo/ VGG & with VGG
\end{tabularx}
\caption{
\textbf{Ablation: VGG loss} improves photo-realism compared to only using a per-pixel L1 loss. \moniker{} is able to render full images efficiently during training, enabling the usage of various image- and patch-based losses.
\label{fig:vgg}
}
\end{center}
\end{figure}

In Tab.~\ref{tab:efficiency} and Fig.~\ref{fig:speed}, we show that \moniker{} is considerably faster to train and render than implicit head avatar methods. Thanks to the coarse-to-fine learning strategy and point pruning, our method only needs to render a small number of points with large radii in early training stages, which significantly speeds up training. 
With the ability to render full images efficiently during training, \moniker{} can be trivially combined with various image- and patch-based training objectives, which target certain properties, \eg, photo-realism, much more effectively than pixel-based losses. In Fig.~\ref{fig:vgg}, we ablate our VGG feature loss, revealing its significant effect in boosting photo-realism.

\subsection{Ablation: Free Model Canonical Space}\label{sec:ablation_canonical}
\begin{figure}
\begin{center}
\newcommand{\myheight}{2.2cm}
\setlength\tabcolsep{1pt}
\begin{tabularx}{\textwidth}{cc}
\includegraphics[height=\myheight]{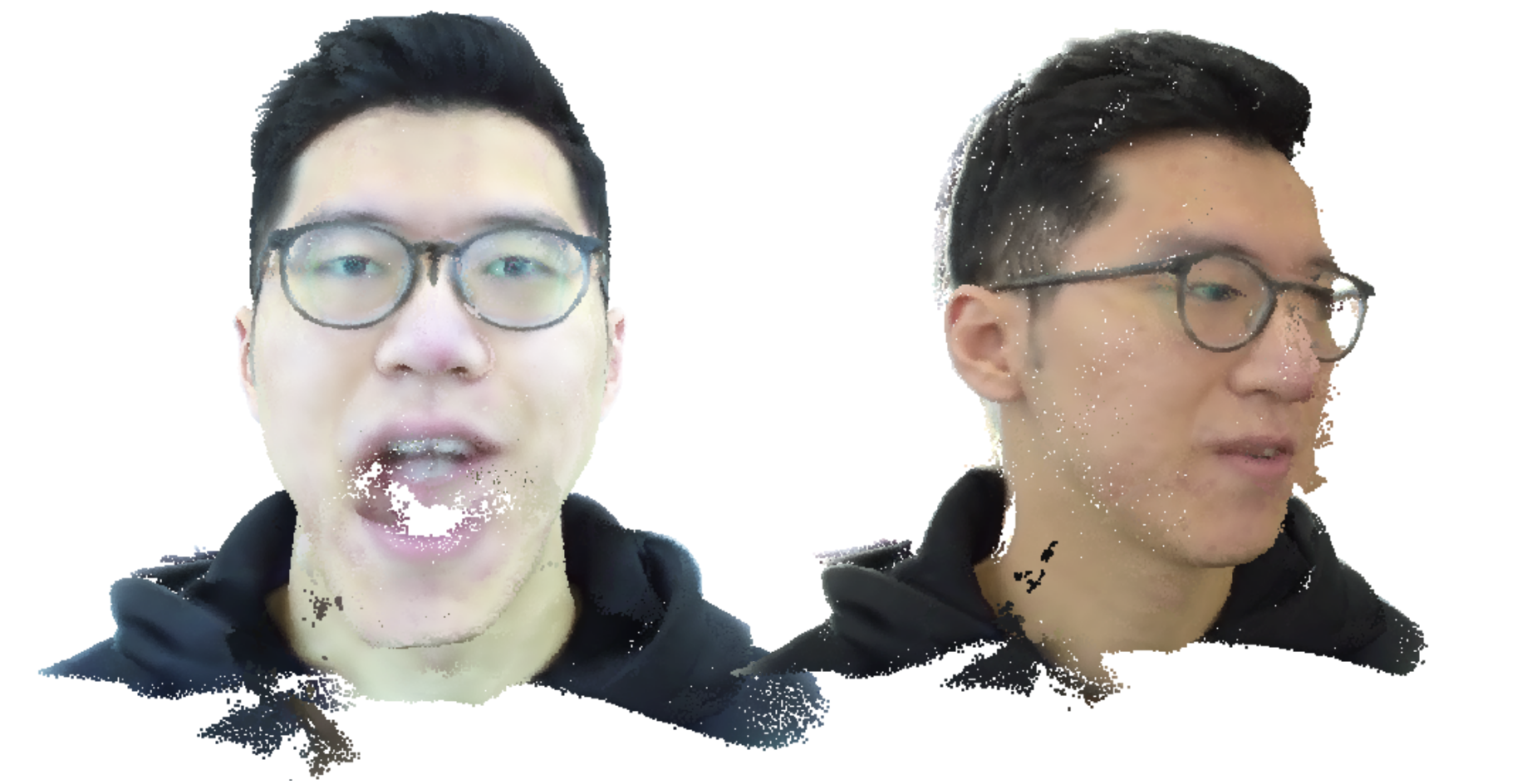}
&\includegraphics[height=\myheight]{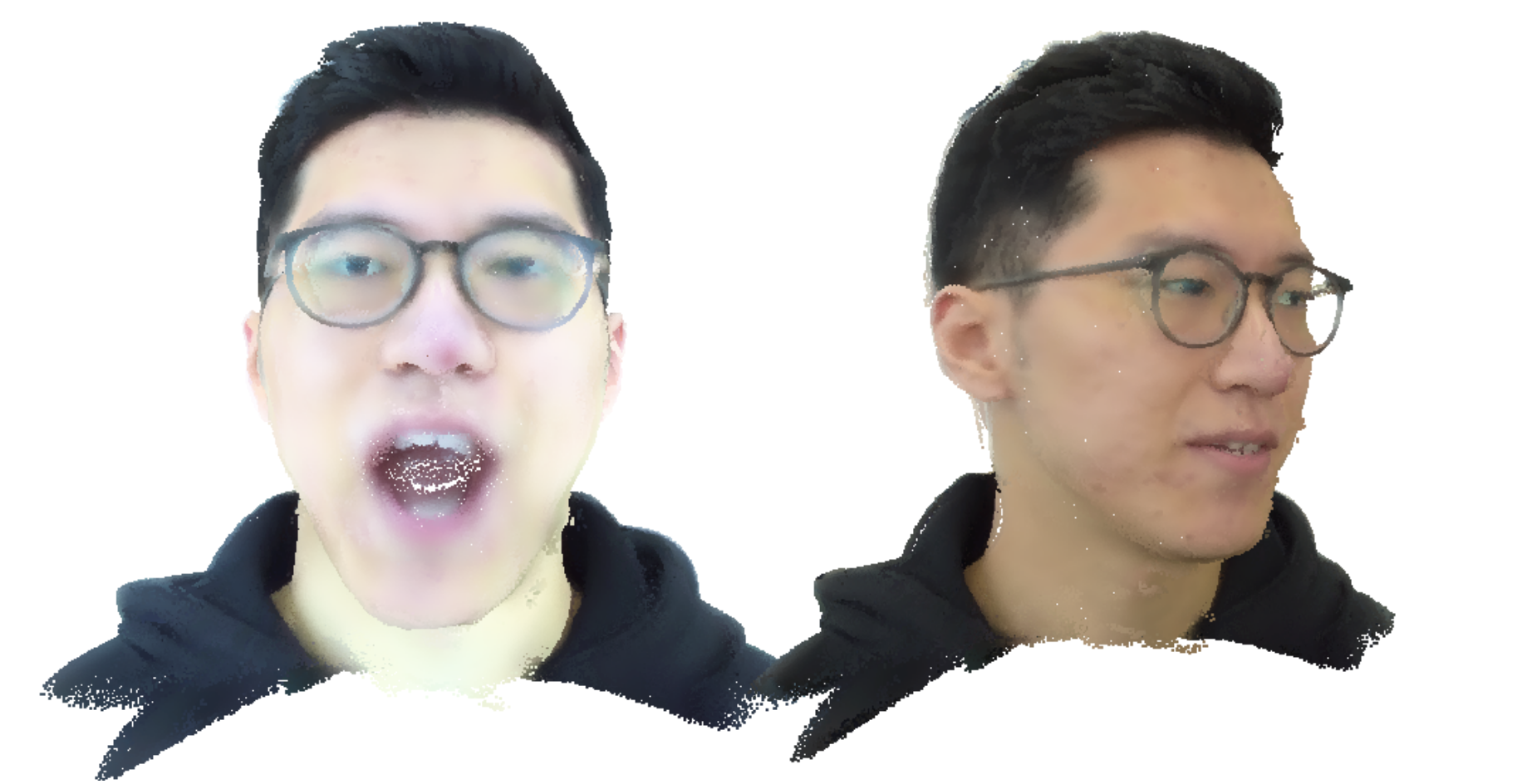}
\\
without canonical offset & with canonical offset
\end{tabularx}
\caption{
\textbf{Ablation: canonical offset} improves the canonical 3D geometry and therefore boosts generalization to novel head poses. For both cases, we show the canonical point representation (transformed to the FLAME canonical space if using canonical offset) and a deformed representation in a novel pose. 
\label{fig:canonical_offset}
}
\end{center}
\end{figure}

As depicted in \cref{fig:canonical_offset}, having an additional freely-learned canonical space, compared to forcing the model to directly learn in a predefined FLAME space, improves the canonical point geometry and generalization to novel poses significantly. 
A possible reason is that the deformations, modeled by an MLP, are easier to optimize compared to point locations. 
Without the canonical offset \(\mathcal{O}\), the model overfits and learns wrong canonical geometries. 

\section{Discussion}
We propose \moniker, a deformable point-based avatar representation that features high flexibility, efficient rendering and straightforward deformation. We show that our method is able to handle various challenging cases in the task of head avatar modeling, including eyeglasses, voluminous hair, skin details and extreme head poses. %
Despite being trained only on a monocular video with fixed lighting conditions, \moniker{} achieves detailed facial and hair geometry and faithfully disentangles lighting effects from the intrinsic albedo. %

There are several exciting future directions. 
\begin{inparaenum}[(1)]
\item Our shading MLP maps normal directions to pose-dependent lighting effects, without further disentangling them into environment maps and surface reflectance properties, limiting relighting capability. Future works could leverage a more constrained physically-based rendering model. 
\item We render points with uniform radius but some regions require denser and finer points to be modeled accurately, \eg, eyes and hair strands. Rendering with different point sizes could potentially achieve detailed reconstructions with less points, and speed up training and rendering further.
\item We focus on the explainability of our method, but future works could combine \moniker{} with 2D neural rendering to boost photo-realism.
\item Our method cannot faithfully model the reflection of eyeglass lenses, which can be improved by modeling transparencies and reflections~\cite{verbin2022refnerf}.
\end{inparaenum}

\noindent\textbf{Acknowledgements}
Yufeng Zheng is supported by the Max Planck ETH Center for Learning Systems. 
Wang Yifan is partially funded by the SNF postdoc mobility fellowship. 
MJB has received research gift funds from Adobe, Intel, Nvidia, Meta/Facebook, and Amazon. MJB has financial interests in Amazon, Datagen Technologies, and Meshcapade GmbH. While MJB is a part-time employee of Meshcapade, his research was performed solely at, and funded solely by, the Max Planck Society.

\begin{center}
\textbf{\large Supplemental Materials}
\end{center}
\setcounter{equation}{0}
\setcounter{figure}{0}
\setcounter{table}{0}
\setcounter{page}{1}
\setcounter{section}{0}

In this supplemental document, we provide the additional experiments mentioned in the main paper in \cref{sec:additional result}, implementation details for our method in \cref{sec:implementation details}, the proof for Jacobian-based normal transformation in \cref{sec:proof}, and discussions about data capture ethics in \cref{sec:capture_ethics}. 
Additionally, we recommend checking out our supplemental video.
\section{Additional Results}
\label{sec:additional result}
\subsection{Results on the MakeHuman Dataset}
\label{sec:makehuman}
\begin{figure}[H]
\begin{center}
\newcommand{\myheight}{2.3cm}
\setlength\tabcolsep{1pt}
\begin{tabularx}{\textwidth}{cccc}
\includegraphics[trim=8em 8em 4em 0em, clip=true, height=\myheight]{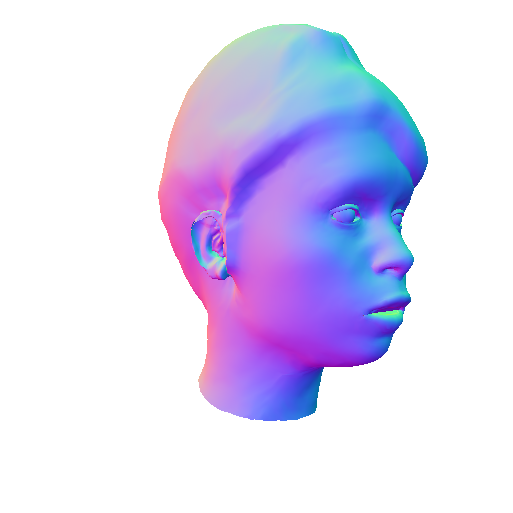}
&\includegraphics[trim=8em 8em 4em 0em, clip=true, height=\myheight]{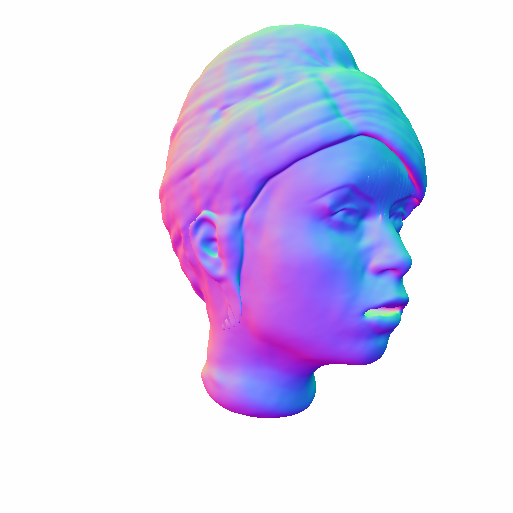}
&\includegraphics[trim=8em 8em 4em 0em, clip=true, height=\myheight]{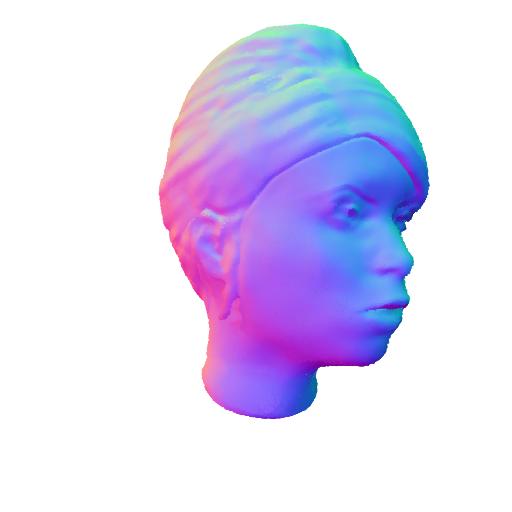}
&\includegraphics[trim=8em 8em 4em 0em, clip=true, height=\myheight]{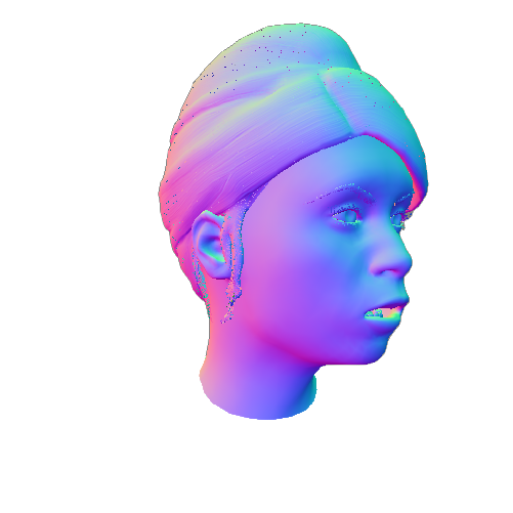}
\\
NHA~\cite{Grassal_2022_nha} & IMavatar~\cite{Zheng_2022_imavatar} & Ours & Ground-Truth
\end{tabularx}
\caption{
\textbf{Qualitative comparison of surface geometry.} Our point-based avatar representation reconstructs comparable head geometry to surface-only representations such as meshes (NHA) and implicit surfaces (IMavatar). %
\label{fig:makehuman}
}
\end{center}
\end{figure}

We evaluate the reconstructed surface geometry on the MakeHuman synthetic dataset rendered from \cite{Briceno_2019_makehuman}. We show that \moniker{} is not only capable of capturing volumetric structures as shown in the main paper, but also performs on par with state-of-the-art methods on surface geometry reconstruction, as shown in \cref{fig:makehuman} and \cref{tab:makehuman}. 

\begin{table}[H]
\resizebox{\linewidth}{!}{
    \begin{tabular}{lcccc}
    \toprule
    Method & female 1 & female 2 & male 1 & male 2\\
    \midrule
    NHA~\cite{Grassal_2022_nha}  & 0.94& 0.95& 0.94& 0.94 \\
    IMavatar~\cite{Zheng_2022_imavatar} & \textbf{0.961} & \textbf{0.966} & \textbf{0.954} & 0.955\\
    Ours &0.954 & 0.954 & 0.944 & \textbf{0.958} \\
    \bottomrule
    \end{tabular}
}
\caption{\textbf{Quantitative comparison on the MakeHuman dataset~\cite{Briceno_2019_makehuman}.} We report the normal consistency metric for evaluating reconstructed surface geometry. The scores for NHA~\cite{Grassal_2022_nha} and IMavatar~\cite{Zheng_2022_imavatar} are obtained from their papers.
}
\label{tab:makehuman}
\end{table}
We thank the authors of NHA~\cite{Grassal_2022_nha} for kindly sharing this evaluation dataset with us. 
\subsection{Learning without Foreground Masks}
\begin{figure}[H]
\begin{center}
\newcommand{\myheight}{2cm}
\setlength\tabcolsep{1pt}
\begin{tabularx}{\textwidth}{cccc}
\includegraphics[height=\myheight]{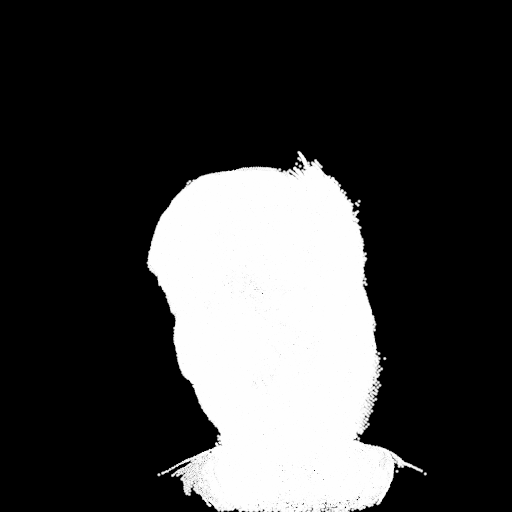}
&\includegraphics[height=\myheight]{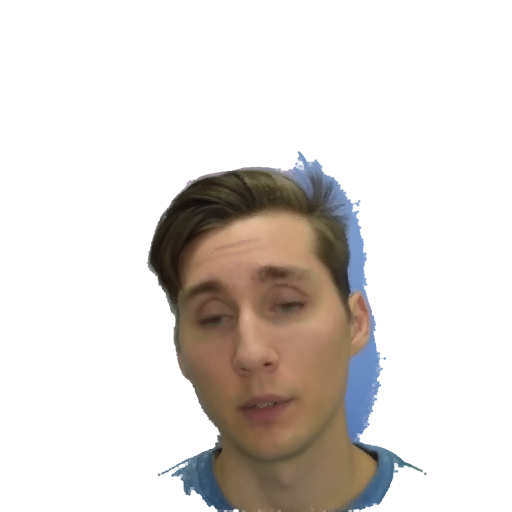}
&\includegraphics[height=\myheight]{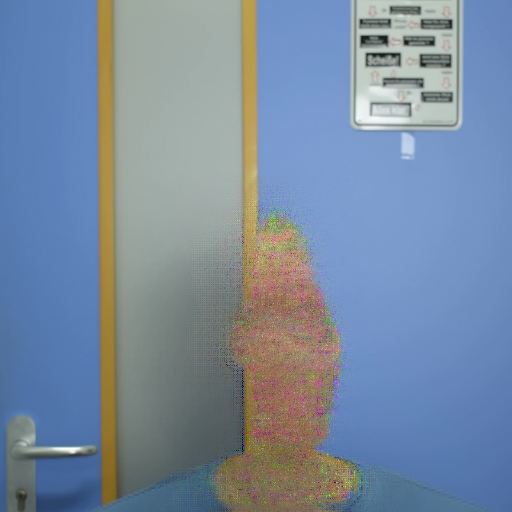}
&\includegraphics[height=\myheight]{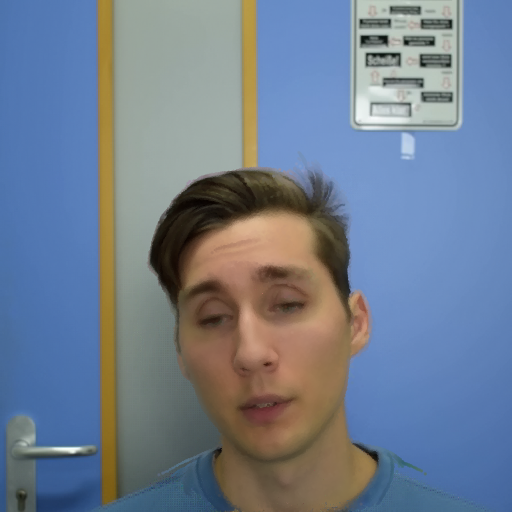}
\\
learned mask & foreground & background & full
\end{tabularx}
\caption{
\textbf{Self-supervised foreground-background disentanglement.} \moniker{} can be learned without mask supervision or known background. 
\label{fig:learned_bkgd}
}
\end{center}
\end{figure}
It is beneficial to learn avatars without relying on foreground segmentation masks, because off-the-shelf segmentation networks often fail in in-the-wild scenarios. In Fig.~\ref{fig:learned_bkgd}, we show that \moniker{} is able to roughly disentangle foreground and background contents in a self-supervised manner. Note that NerFace~\cite{Gafni_2021_nerface} cannot be learned without known backgrounds.

\section{Implementation Details}
In this section, we provide implementation details regarding network architectures, training and evaluation procedures, and the preprocessing of videos.
\label{sec:implementation details}
\subsection{Network Architecture}
\begin{figure}[t]
\begin{center}
\includegraphics[trim=0em 0em 0em 0em, clip=true, width=\linewidth]{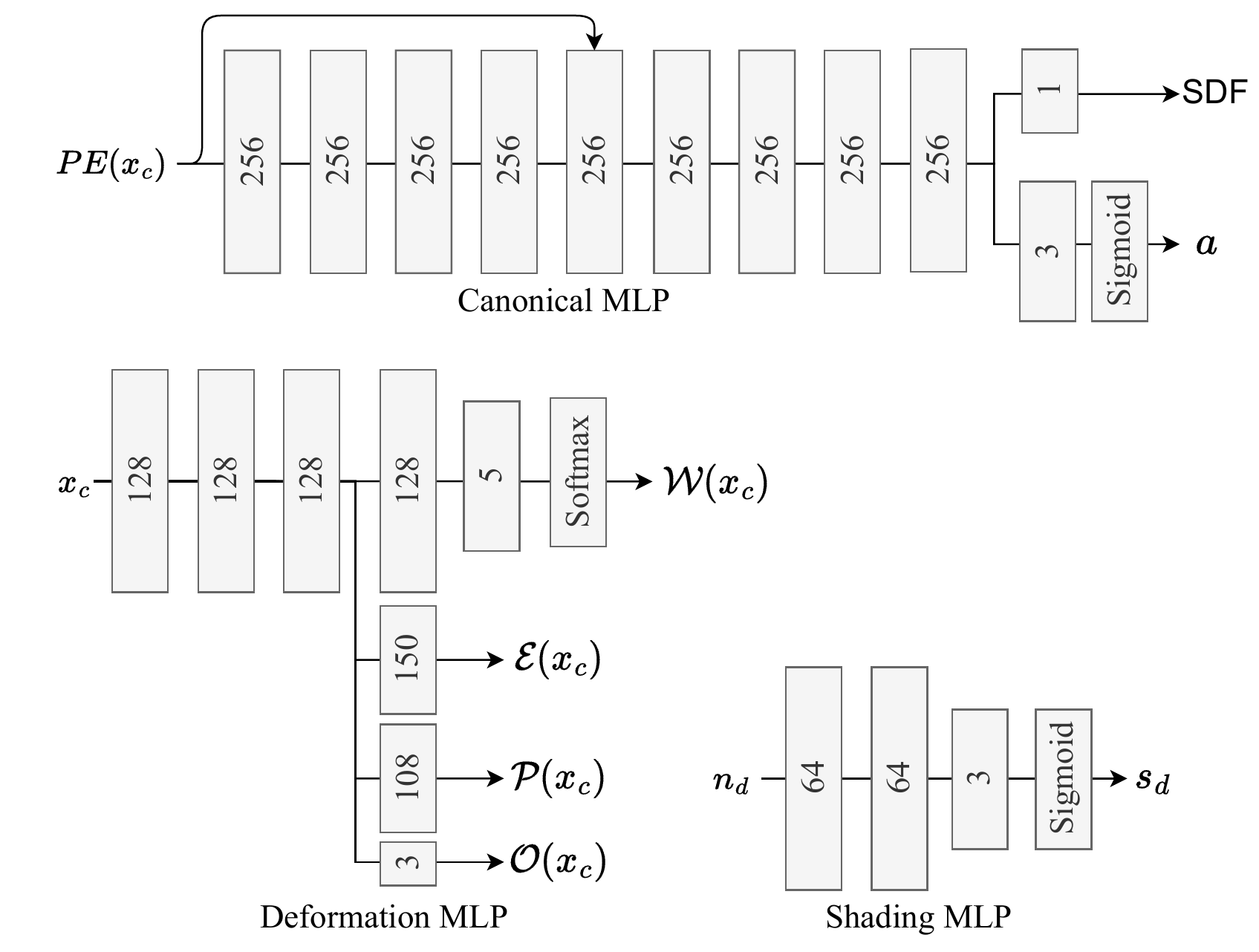}
\end{center}
\caption{\textbf{Network architecture. }We show the network architectures of the canonical, deformation and shading MLPs. Except from the last layer, each linear layer is followed by weight normalization~\cite{Salimans16} and non-linear activation. We use the Softplus~\cite{Dugas00} function for the canonical and deformation MLP, and the ReLU activation for the shading MLP. }
\label{fig:architecture}
\end{figure}

We show the architecture of the canonical, deformation and shading MLPs in \cref{fig:architecture}. 
\subsection{Training Details}
\label{sec:traindetails}
\paragraph{Loss weights and optimization.}We choose \(\lambda_{\textrm{rgb}} = 1 \textrm{, } \lambda_{\textrm{mask}} = 1\textrm{, } \lambda_{\textrm{flame}} = 1\textrm{, and } \lambda_{\textrm{vgg}} = 0.1\) for most of our experiments, except for the experiment shown in Fig.~6 (main paper), where we used \(\lambda_{\textrm{rgb}} = 5 \textrm{, and } \lambda_{\textrm{vgg}} = 0.05\), and the experiment on synthetic data (\cref{sec:makehuman} in \suppmat), where we used \(\lambda_{\textrm{rgb}} = 10 \textrm{, and } \lambda_{\textrm{vgg}} = 0\). For the flame loss, we choose \(\lambda_{\textrm{e}} = 1000 \textrm{, } \lambda_{\textrm{p}}=1000 \textrm{, and } \lambda_{\textrm{w}}=1\). For optimizing the canonical SDF, we use \(\lambda_{\textrm{SDF}} =1\textrm{, and } \lambda_{\textrm{eik}}=0.1\). We train our models using an Adam optimizer~\cite{kingma14} for 63 epochs, with a learning rate of $\eta=1e^{-4}$, and $\beta = (0.9, 0.999)$. 
\paragraph{Static bone transformation. } In the pre-processing step, the FLAME tracking of the upper body is often unstable.
We handle this problem via the introduction of an additional static bone, similar to IMavatar~\cite{Zheng_2022_imavatar}. Since the upper body remains mostly static in the training video, the deformation MLP assigns large LBS weights to the static bone to reduce shoulder motion under different FLAME pose configurations. This trick effectively reduces shoulder jittering. 
\paragraph{Point upsampling and pruning.} Every \(5\) epochs, we double the number of points and reduce the point rasterization radii, which allows our method to capture finer details. The new point locations are obtained by perturbating the current points with random noises. The factor for radius reduction is carefully chosen: Assuming that points roughly form a surface, the radius should be decreased by a factor of \(1/\sqrt{2}\). We choose to reduce the point radii by a factor of \(0.75\) in practice. Additionally, after each epoch, we prune away points that are invisible, \ie, not projected onto any pixels with compositing weights larger than a threshold. We choose \(0.5\) as the pruning threshold.

\subsection{Evaluation Details}
\begin{figure}[H]
\vskip -2mm
\begin{center}
\newcommand{\myheight}{3.0cm}
\setlength\tabcolsep{1pt}
\begin{tabularx}{\textwidth}{cccc}
\includegraphics[trim=0.9em 0em 0.4em 0em, clip=true, height=\myheight]{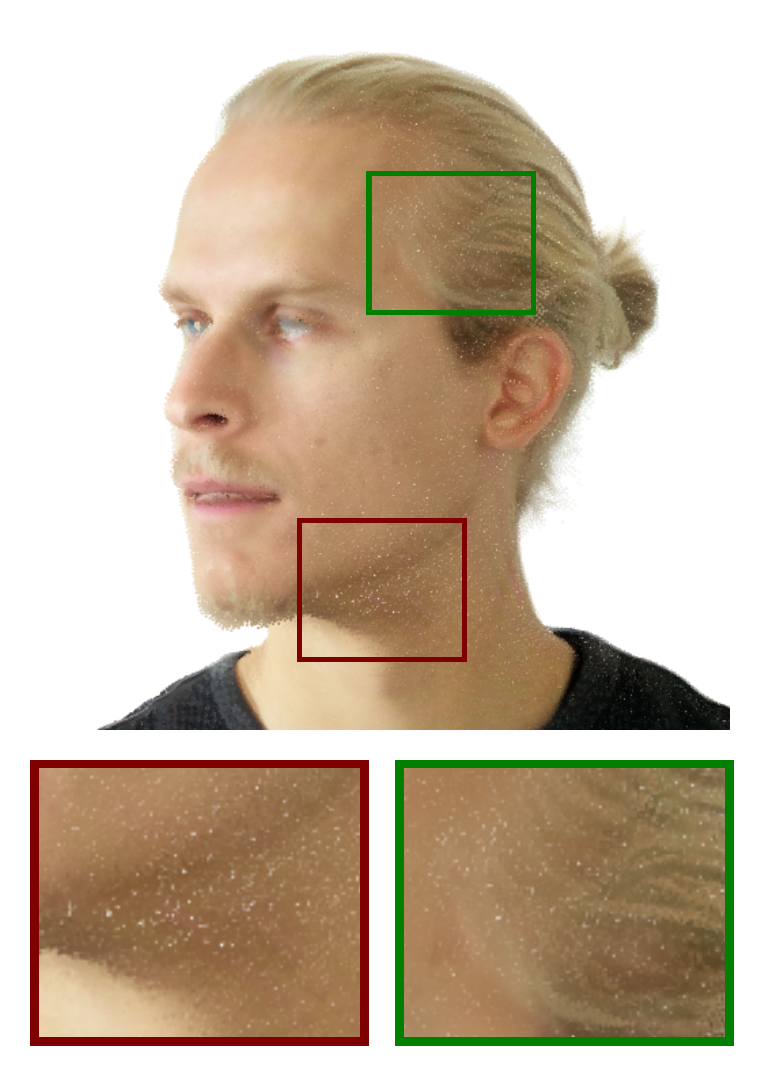}
&\includegraphics[trim=0.9em 0em 0.4em 0em, clip=true, height=\myheight]{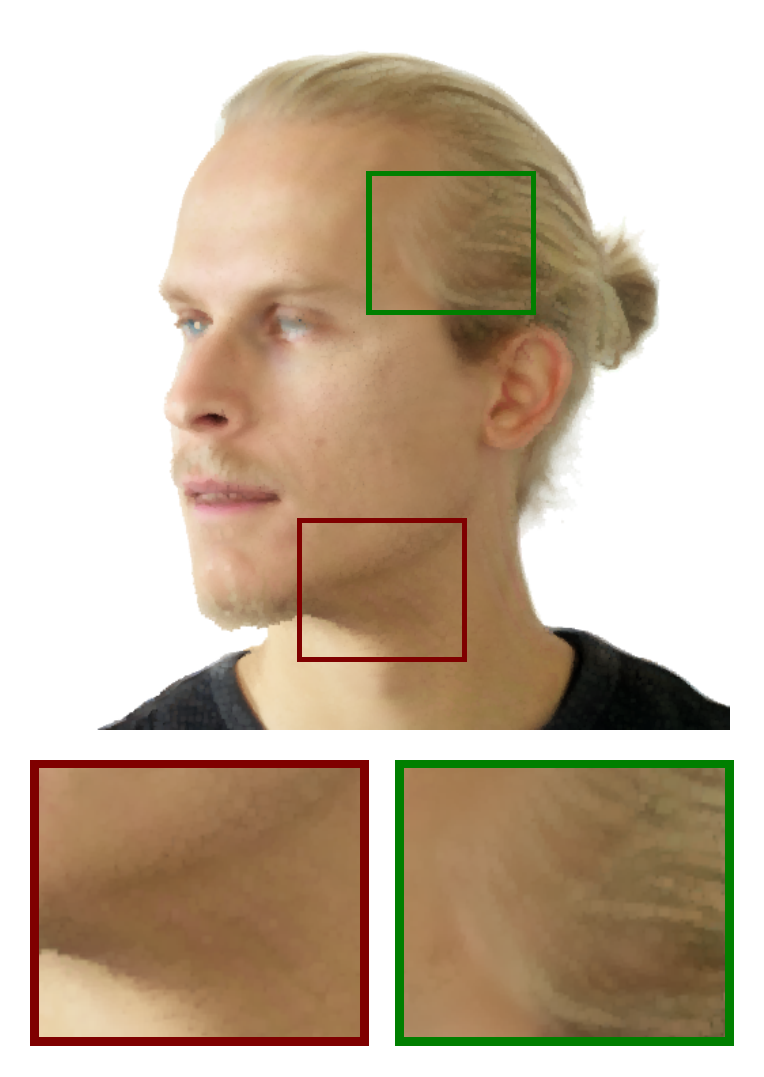}
&\includegraphics[trim=0.9em 0em 0.4em 0em, clip=true, height=\myheight]{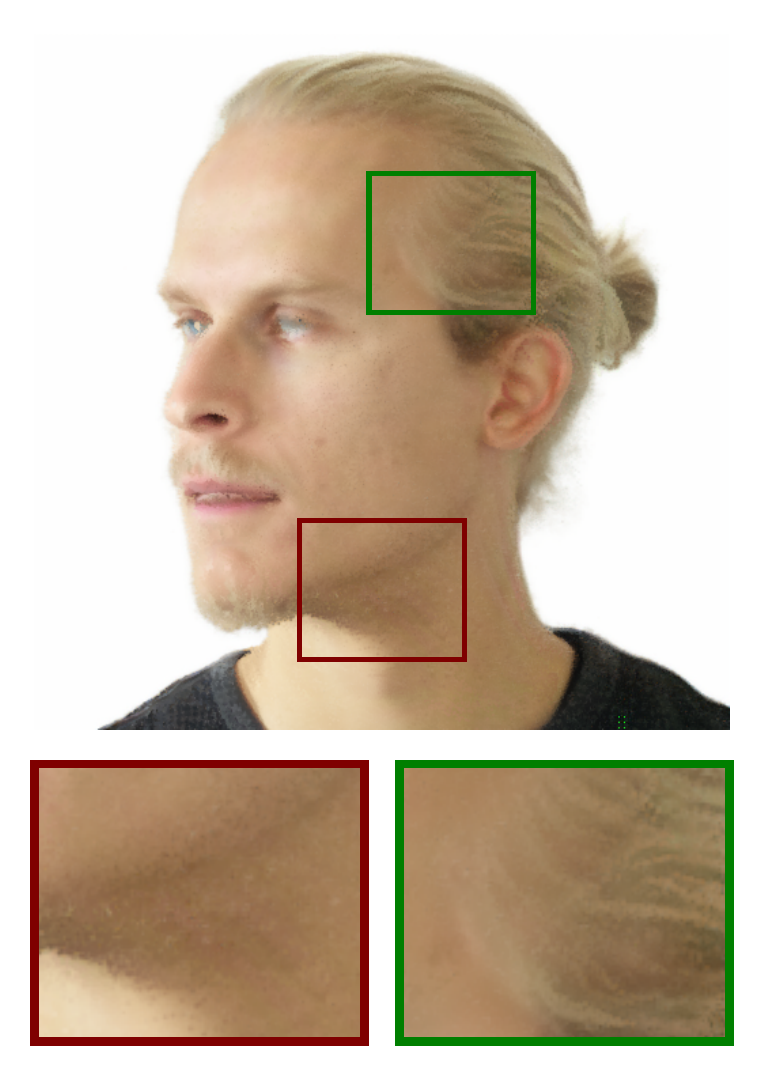}
&\includegraphics[trim=0.9em 0em 0.9em 0em, clip=true, height=\myheight]{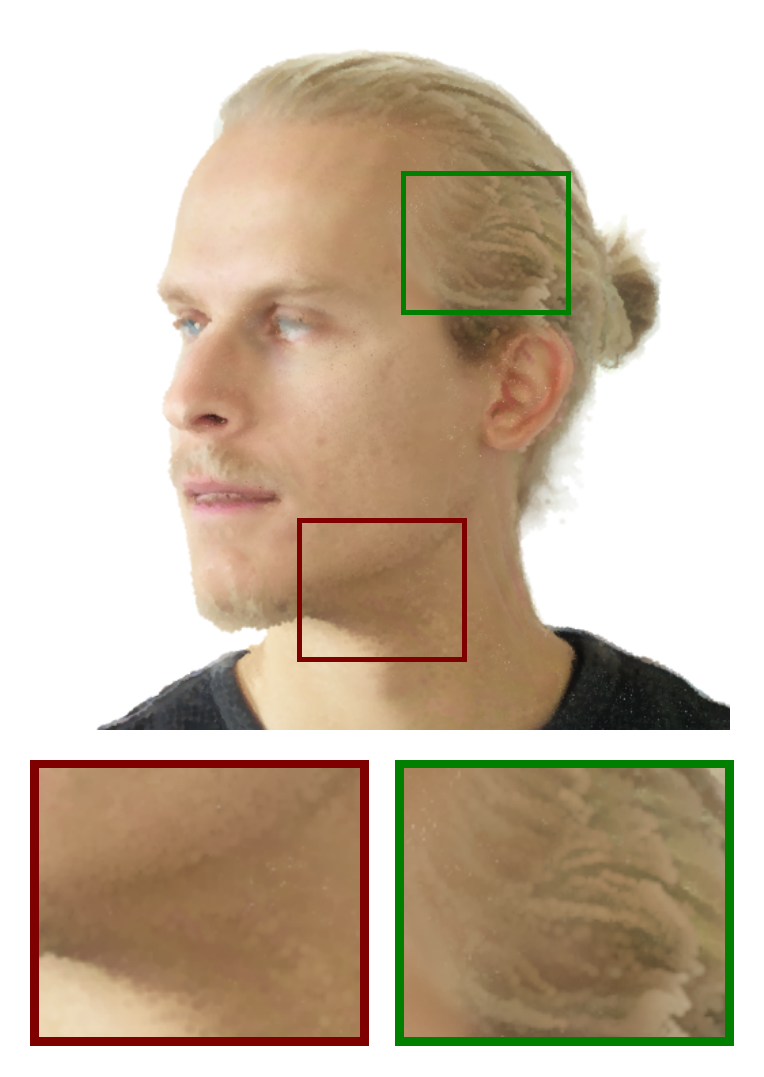}
\\
orig. RGB & (1) erode-dilate & (2) 2D NN & (3) adapt radii
\end{tabularx}
\caption{
\textbf{Point sparsity and white dot artifacts.} We show the rendered image which contains white dot artifacts (orig. RGB), and refined results from different post-processing methods. 
\label{fig:sparsity}
}
\end{center}
\vskip -5mm
\end{figure}

\paragraph{FLAME~\cite{Li_2017_flame} parameter optimization.}Similar to NHA~\cite{Grassal_2022_nha}, our method also fine-tunes pre-tracked FLAME expression and pose parameters during training and evaluation. For training, the loss weights are elaborated in the main paper and in \cref{sec:traindetails} of \suppmat{} For test-time tracking optimization, we only use the RGB loss. 
\paragraph{Hole filling.}%
Rendered images from \moniker{} may contain ``white dot" artifacts. This happens due to the sparsity of deformed point clouds which reveals the white background. The issue mainly occurs around 
\begin{inparaenum}[(1)]
\item stretched regions (\eg neck), in this case, the deformation causes sparsity;
\item regions that are infrequently observed during training (\eg profile views), because unseen points are pruned during training (see Sec.~2.2 of \suppmat~for pruning strategy).
\end{inparaenum}
In Fig.~\ref{fig:sparsity}, we compare several post-processing methods to address this artifact. 
\begin{inparaenum}[(1)]
\item We apply erosion and dilation consecutively to the rendered images. This method fills most white dots but slightly blurs images. 
\item We train a per-subject 2D neural network (NN) to remove white dots from the rendered images. This method robustly removes artifacts and does not blur images, but it requires cumbersome per-subject NN training. More specifically, we augment GT images with white dots using the rendered mask, and train a 2D NN to remove the added artifacts. We found the trained NN can remove artifacts from rendered images despite the domain gap, and also generalizes to unseen poses. 
\item We increase point radii based on distances to neighbors before rasterization. This method principally tackles the sparsity problem in 3D space, but often falsely thickens thin structures such as hair strands.
\end{inparaenum}
We use the first method in the paper due to its simplicity and effectiveness, but adaptive radii and neural networks could also be leveraged for better performance in some cases.

\paragraph{Relighting.}%
Our method learns to disentangle \emph{albedo} from pose-dependent lighting effects which we refer to as \emph{shading}. 
Previous works that use MLP-based texture networks transform 3D locations and normal directions into shaded colors. In contrast, we perform disentanglement by conditioning the albedo only on the canonical locations and the shading only on the normal directions. This is how we achieve \emph{unsupervised} albedo disentanglement. 

We found our method allows for rudimentary relighting
\begin{inparaenum}[(1)]
\item by manipulating the normal directions of the shading component, which changes the light direction (please check the teaser and Fig.~5 in the main paper for horizontally-flipped and rotated light directions), or 
\item by discarding the learned shading and instead rendering with a default shading model using the learned albedo and normals. 
\end{inparaenum}
The latter option allows for relighting in novel scenes. In the supplementary video (03:39), we use a diffused shading model to relight avatars, but it leads to less faithful appearances compared to the learned shading.

\subsection{Data Preprocessing}
We leverage the preprocessing pipeline of IMavatar~\cite{Zheng_2022_imavatar} and use the same camera and FLAME parameters for all SOTA methods. This allows us to compare head avatar methods without considering the influence of different face-tracking schemes used for preprocessing.
\section{Proof of Eq. 6 (deformed normal formula)}
\label{sec:proof}
First, we recap Eq. 6 from the main paper, which states that deformed normals can be obtained as:
\begin{equation*}
\mathbf{n_d} = l\mathbf{n_c}\left(\frac{\mathrm{d} \mathbf{x_d}}{\mathrm{d} \mathbf{x_c}}\right)^{-1}
\end{equation*}
To prove it, we assume points form flat surfaces locally. The deformed normal \(\mathbf{n_d}\) of a point can then be defined via the constraint:
\begin{equation*}
\mathbf{n_d} \cdot (\mathbf{x_{d,i}} - \mathbf{x_d}) = 0, 
\end{equation*}
which needs to be satisfied for every neighboring point \(\mathbf{x_{d,i}}\). Note that \(\mathbf{x_{d,i}}\) can be linearly approximated as 
\begin{equation*}
\mathbf{x_{d,i}} = \mathbf{x_d} +  \frac{\mathrm{d} \mathbf{x_d}}{\mathrm{d} \mathbf{x_c}} (\mathbf{x_{c,i}} - \mathbf{x_c}),
\end{equation*}
which allows us to rewrite the constraint as 
\begin{equation*}
\mathbf{n_d} \cdot \frac{\mathrm{d} \mathbf{x_d}}{\mathrm{d} \mathbf{x_c}} (\mathbf{x_{c,i}} - \mathbf{x_c}) = 0.
\end{equation*}
Note that the canonical normal \(\mathbf{n_c}\) satisfies 
\begin{equation*}
\mathbf{n_c} \cdot (\mathbf{x_{c,i}} - \mathbf{x_c}) = 0.
\end{equation*}
Therefore, in order to satisfy the constraint for all neighboring points \(\mathbf{x_{d,i}}\), \(\mathbf{n_d}\) must satisfy 
\begin{equation*}
\mathbf{n_d} \cdot \frac{\mathrm{d} \mathbf{x_d}}{\mathrm{d} \mathbf{x_c}}  = l \mathbf{n_c},
\end{equation*}
where \(l\) is a normalizing scalar to ensure unit normal length. This leads to the formulation in Eq. 6.  
\section{Ethics}
\label{sec:capture_ethics}
We captured 5 human subjects with smartphones or laptop cameras for our experiments. All subjects have given written consents for using the captured images in this project. We will make the data publicly available for research purposes, where permission for data publishing is given by the subjects. 

Our method could be extended to generate media content of real people performing synthetic poses and expressions. We do not condone using our work to generate fake images or videos of any person with the intent of spreading misinformation or tarnishing their reputation.

{\small
\bibliographystyle{ieee_fullname}
\bibliography{egbib}
}

\end{document}